# SCREEN: Learning a Flat Syntactic and Semantic Spoken Language Analysis Using Artificial Neural Networks

**Stefan Wermter**                    WERMTER@INFORMATIK.UNI-HAMBURG.DE
**Volker Weber**                      WEBER@INFORMATIK.UNI-HAMBURG.DE
*Department of Computer Science*
*University of Hamburg*
*22527 Hamburg, Germany*

## Abstract

Previous approaches of analyzing spontaneously spoken language often have been based on encoding syntactic and semantic knowledge manually and symbolically. While there has been some progress using statistical or connectionist language models, many current spoken-language systems still use a relatively brittle, hand-coded symbolic grammar or symbolic semantic component.

In contrast, we describe a so-called screening approach for *learning robust processing* of spontaneously spoken language. A screening approach is a flat analysis which uses shallow sequences of category representations for analyzing an utterance at various syntactic, semantic and dialog levels. Rather than using a deeply structured symbolic analysis, we use a flat connectionist analysis. This screening approach aims at supporting speech and language processing by using (1) data-driven learning and (2) robustness of connectionist networks. In order to test this approach, we have developed the SCREEN system which is based on this new robust, learned and flat analysis.

In this paper, we focus on a detailed description of SCREEN's architecture, the flat syntactic and semantic analysis, the interaction with a speech recognizer, and a detailed evaluation analysis of the robustness under the influence of noisy or incomplete input. The main result of this paper is that flat representations allow more robust processing of spontaneous spoken language than deeply structured representations. In particular, we show how the fault-tolerance and learning capability of connectionist networks can support a flat analysis for providing more robust spoken-language processing within an overall hybrid symbolic/connectionist framework.

## 1. Introduction

Recently the fields of speech processing as well as language processing have both seen efforts to examine the possibility of integrating speech and language processing (von Hahn & Pyka, 1992; Jurafsky et al., 1994b; Waibel et al., 1992; Ward, 1994; Menzel, 1994; Geutner et al., 1996; Wermter et al., 1996). While new and large speech and language corpora are being developed rapidly, new techniques have to be examined which particularly support properties of both speech and language processing. Although there have been quite a few approaches to spoken-language analysis (Mellish, 1989; Young et al., 1989; Hauenstein & Weber, 1994; Ward, 1994), they have not emphasized *learning* a syntactic and semantic analysis of spoken language using a hybrid connectionist[1] architecture which is the topic of this paper and our goal in SCREEN[2]. However, learning is important for the reduction of knowledge acquisition, for automatic system adaptation, and for increasing the system portability for new domains. Different from most previous approaches, in this paper we demonstrate that hybrid connectionist learning techniques can be used for providing a robust flat analysis of faulty spoken language.

Processing spoken language is very different from processing written language, and successful techniques for text processing may not be useful for spoken-language processing. Processing spoken language is less constrained, contains more errors and less strict regularities than written language. Errors occur on all levels of spoken-language processing. For instance, acoustic errors, repetitions, false starts and repairs are prominent in spontaneously spoken language. Furthermore, incorrectly analyzed words, unforeseen grammatical and semantic constructions occur very often in spoken language. In order to deal with these important problems for "real-world" language analysis, robust processing is necessary. Therefore we cannot expect that existing techniques like context-free tree representations which have been proven to work for written language can simply be transferred to spoken language.

For instance, consider that a speech recognizer has produced the correct German sentence hypothesis "Ich meine natürlich März" (English translation: "I mean of_course March"). Standard techniques from text processing - like chart parsers and context-free grammars - may be able to produce deeply structured tree representations for many correct sentences as shown in Figure 1.

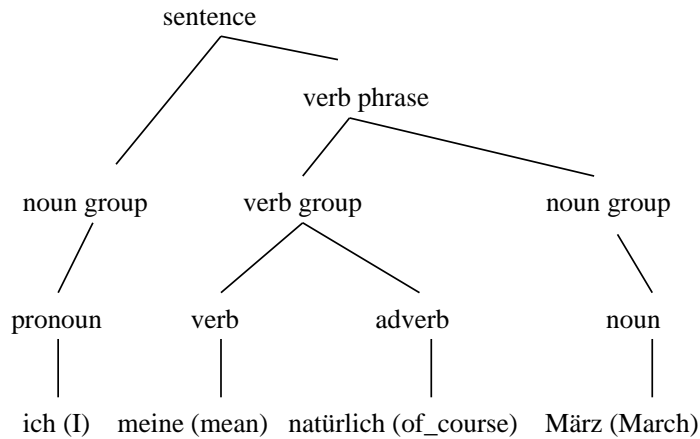

Figure 1: Tree representation for a correctly recognized sentence

However, currently speech recognizers are still far from perfect and produce many word errors so that it is not possible to rely on a perfect sentence hypothesis. Therefore, incorrect

---

1. Sometimes connectionist networks are also called artificial neural networks. From now on we will use only the term "connectionist networks", and the term "hybrid connectionist architecture" to refer to an architecture which emphasizes the use of connectionist networks but does not rule out the use of symbolic representations on higher levels where they might be needed.

2. **S**ymbolic **C**onnectionist **R**obust **E**nterpris**E** for **N**atural language

variations like "Ich meine ich März" ("I mean I March"), "Ich hätte ich März" ("I had I March") and "Ich Ich meine März" ("I I mean March") have to be analyzed. However, in context-free grammars a single syntactic or semantic category error may prevent that a complete tree can be built, and standard top-down chart parsers may fail completely. However, suboptimal sentence hypotheses have to be analyzed since sometimes such sentence hypotheses are the best possible output produced by a speech recognizer. Furthermore, a lot of the content can be extracted even from partially incorrect sentence hypotheses. For instance, from "I had I March" it is plausible that an agent "I" said something about the time "March". Therefore, a robust analysis should be able to analyze such sentence hypotheses and ideally should not break for any input.

## 1.1 Screening Approach: Flat Representations Support Robustness

For such examples of incorrect variations of sentence hypotheses, an in-depth structured syntactic and semantic representation is not advantageous since more arbitrary word order and spontaneous errors make it often impossible to determine a desired deep highly structured representation. Furthermore, a deep highly structured representation may have many more restrictions than appropriate for spontaneously spoken language. However, and maybe even more important, for certain tasks it is not necessary to perform an in-depth analysis. While, for instance, inferences about story understanding require an in-depth understanding (Dyer, 1983), tasks like information extraction from spoken language do not need much of an in-depth analysis. For instance, if the output of our parser were to be used for translating a speech recognizer sentence hypothesis "Eh ich meine eh ich März" ("Eh I mean eh I March"), it may be sufficient to extract that an agent ("I") uttered ("mean") a time ("March"). In contrast to a deeply structured representation, our screening approach aims at reaching a flat but robust representation of spoken language. A *screening approach* is a shallow flat analysis based on category sequences (called flat representations) at various syntactic and semantic levels.

A *flat representation* structures an utterance U with words $w_1$ to $w_n$ according to the syntactic and semantic properties of the words in their contexts, e.g., according to a sequence of basic or abstract syntactic categories. For instance, the phrase "a meeting in London" can be described as a flat representation "determiner noun preposition noun" at a basic syntactic level and as a flat representation "noun-group noun-group prepositional-group prepositional-group" at an abstract syntactic level. Similar flat representations can be used for semantic categories, dialog act categories, etc.

| Käse (Rubbish) | ich (I) | meine (mean) | natürlich (of_course) | März (March) |
|---|---|---|---|---|
| noun | pronoun | verb | adverb | noun |
| no | animate | utter | nil | time |
| noun group | noun group | verb group | special group | noun group |
| negation | agent | action | miscellaneous | at time |

Figure 2: Utterance with its flat representation

Figure 2 gives an example for a flat representation for a correct sentence hypothesis "Käse ich meine natürlich März" ("Rubbish I mean of course March"). The first line shows the sentence, the second its literal translation. The third line describes the basic syntactic category of each word, the fourth line shows the basic semantic category. The last two lines illustrate the syntactic and semantic categories at the phrase level.

| Käse | ich | hätte | ich | März |
|------|-----|-------|-----|------|
| (Rubbish) | (I) | (had) | (I) | (March) |
| noun | pronoun | verb | pronoun | noun |
| no | animate | have | animate | time |
| noun group | noun group | verb group | noun group | noun group |
| negation | agent | action | agent | at time |

Figure 3: Utterance with its flat representation

Figure 3 gives an example for a flat representation for the incorrect sentence hypothesis "Käse ich hätte ich März" ("Rubbish I had I March"). A parser for spoken language should be able to process such sentence hypotheses as far as possible, and we use flat representations to support the necessary robustness. In our example, the analysis should at least provide that an animate agent and noun group ("I") made some statement about a specific time and noun group ("March"). Flat representations have the potential to support robustness better since they have only a minimal sequential structure, and even if an error occurs the whole representation can still be built. In contrast, in standard tree-structured representations many more decisions have to be made to construct a deeply structured representation, and therefore there are more possibilities to make incorrect decisions, in particular with noisy spontaneously spoken language. So we chose flat representations rather than highly structured representations because of the desired robustness against mistakes in speech/language systems.

## 1.2 Flat Representations Learned in a Hybrid Connectionist Framework

Robust spoken-language analysis using flat representations could be pursued in different approaches. Therefore we want to motivate why we use a hybrid connectionist approach, which uses connectionist networks as far as possible but does not rule out the use of symbolic knowledge. So why do we use connectionist networks?

Most important, due to their distributed fault tolerance, connectionist networks support robustness (Rumelhart et al., 1986; Sun, 1994) but connectionist networks also have a number of other properties which are relevant for our spoken-language analysis. For instance, connectionist networks are well known for their learning and generalization capabilities. Learning capabilities allow to induce regularities directly from examples. If the training examples are representative for the task, the noisy robust processing should be supported by inductive connectionist learning.

Furthermore, a hybrid connectionist architecture has the property that different knowledge sources can take advantage of the learning and generalization capabilities of connectionist networks. On the other hand, other knowledge - task or control knowledge - for which rules are known can be represented directly in symbolic representations. Since humans apparently do symbolic inferencing based on real neural networks, abstract models as symbolic representations and connectionist networks have the additional potential to shed some light on human language processing capabilities. In this respect, our approach also differs from other candidates for robust processing, like statistical taggers or statistical n-grams. These statistical techniques can be used for robust analysis (Charniak, 1993) but statistical techniques like n-grams do not relate to the human cognitive language capabilities while simple recurrent connectionist networks have more relationships to the human cognitive language capabilities (Elman, 1990).

SCREEN is a new hybrid connectionist system developed for the examination of flat syntactic and semantic analysis of spoken language. In earlier work we have explored a flat scanning understanding for written texts (Wermter, 1995; Wermter & Löchel, 1994; Wermter & Peters, 1994). Based on this experience we started a completely new project SCREEN to explore a learned fault-tolerant flat analysis for spontaneously spoken-language processing. After preliminary successful case studies with transcripts we have developed the SCREEN system for using knowledge generated from a speech recognizer. In previous work, we gave a brief summary of SCREEN with a specific focus on segmentation parsing and dialog act processing (Wermter & Weber, 1996a). In this paper, we focus on a detailed description of SCREEN's architecture, the flat syntactic and semantic analysis, the interaction with a speech recognizer, and a detailed evaluation analysis of the robustness under the influence of noisy or incomplete input.

## 1.3 Organization and Claim of the Paper

The paper is structured as follows. In Section 2 we provide a more detailed description of examples of noise in spoken language. Noise can be introduced by the human speaker but also by the speech recognizer. Noise in spoken-language analysis motivates the flat representations whose categories are described in Section 3. All basic and abstract categories at the syntactic and semantic level are explained in this section. In Section 4 we motivate and explain the design of the SCREEN architecture. After a brief functional overview, we show the overall architecture and explain details of individual modules up to the connectionist network level. In order to demonstrate the behavior of this flat analysis of spoken language we provide various detailed examples in Section 5. Using several representative sentences we walk the reader through a detailed step-by-step analysis. After the behavior of the system has been explained, we provide the overall analysis of the SCREEN system in Section 6. We evaluate the system's individual networks, compare the performance of simple recurrent networks with statistical n-gram techniques, and show that simple recurrent networks performed better than 1-5 grams for syntactic and semantic prediction. Furthermore we provide an overall system evaluation, examine the overall performance under the influence of additional noise, and supply results from a transfer to a different second domain. Finally we compare our approach to other approaches and conclude that flat representations based on connectionist networks provide a robust learned spoken-language analysis.

We want to point out that this paper does not make an argument against deeply structured symbolic representations for language processing *in general*. Usually, *if* a deeply structured representation can be built, of course due to the additional knowledge it con-tains, its potential for more powerful relationships and interpretations will be greater than that of a flat representation. For instance, in-depth analysis is required for tasks like making detailed planning inferences while reading text stories. However, our screening approach is motivated based on *noisy* spoken-language analysis. For noisy spoken-language analysis, flat representations support robustness, and connectionist networks are effective for providing such robustness due to their learned fault-tolerance. This is a main contribution of our paper, and we demonstrate this by building and evaluating a computational hybrid connectionist architecture SCREEN based on flat, robust, and learned processing.

## 2. Processing Spoken Language

Our goal is to learn to process spontaneously spoken language at a syntactic and semantic level in a fault-tolerant manner. In this section we will give motivating examples of spoken language.

### 2.1 "Noise" in Spoken Language

Our domain in this paper is the arrangement of meetings between business partners, and we currently use 184 spoken dialog turns with 314 utterances from this domain. One turn consists of one or more subsequent utterances of the same speaker. For these 314 utterances, thousands of utterance hypotheses can be generated and have to be processed based on the underlying speech recognizer. German utterance examples from this domain are shown below together with their literal English translation. It is important to note that the English translations are *word-for-word* translations.

1. Käse ich meine natürlich März
   (Rubbish I mean of course March)

2. Der vierzehnte ist ein Mittwoch richtig
   (The fourteenth is a Wednesday right)

3. Ähm am sechsten April bin ich leider außer Hause
   (Eh on sixth April am I unfortunately out of home)

4. Also ich dachte noch in der nächsten Woche auf jeden Fall noch im April
   (So I thought still in the next week in any case still in April)

5. Gut prima vielen Dank dann ist das ja kein Problem
   (Good great many thanks then is this yeah no problem)

6. Oh das ist schlecht da habe ich um vierzehn Uhr dreißig einen Termin beim Zahnarzt
   (Oh that is bad there have I at fourteen o'clock thirty a date at dentist)

7. Ja genau allerdings habe ich da von neun bis vier Uhr schon einen Arzttermin
   (Yes exactly however have I there from nine to four o'clock already a doctor-appointment)

As we can see, spoken language contains many performance phenomena, among them exclamations ("rubbish", see Example 1), interjections ("eh", "so", "oh", see Examples 3, 4 and 6), new starts ("there have I ...", see Example 6). Furthermore, the syntactic and semantic constraints in spoken language are less strict than in written text. For instance, the word order in spontaneously spoken language is often very different from written language. Therefore, spoken language is "noisier" than written language even for these transcribed sentences, and well-known parsing strategies from text processing - which can rely more on wellformedness criteria - are not directly applicable for analyzing spoken language.

## 2.2 "Noise" from a Speech Recognizer

If we want to analyze spoken language in a computational model, there is not only the "noise" introduced by humans while speaking but also the "noise" introduced by the limitations of speech recognizers. Typical speech recognizers produce many separated word hypotheses with different plausibilities over time based on a given speech signal. Such word hypotheses can be connected to a word hypothesis sequence and have to be evaluated for providing a basis for further analysis. Typically, a word hypothesis consists of four parts: 1) the start time in seconds, 2) the end time in seconds, 3) the word string of the hypothesis, and 4) a plausibility of the hypothesis based on the confidence of the speech recognizer. Below we show a simple word graph[3]. In practice, word graphs for spontaneous speech can be much longer leading to comprehensive word hypothesis sequences. However, for illustrating the properties of the speech input we focus on this relatively short and simple word graph (Figure 4).

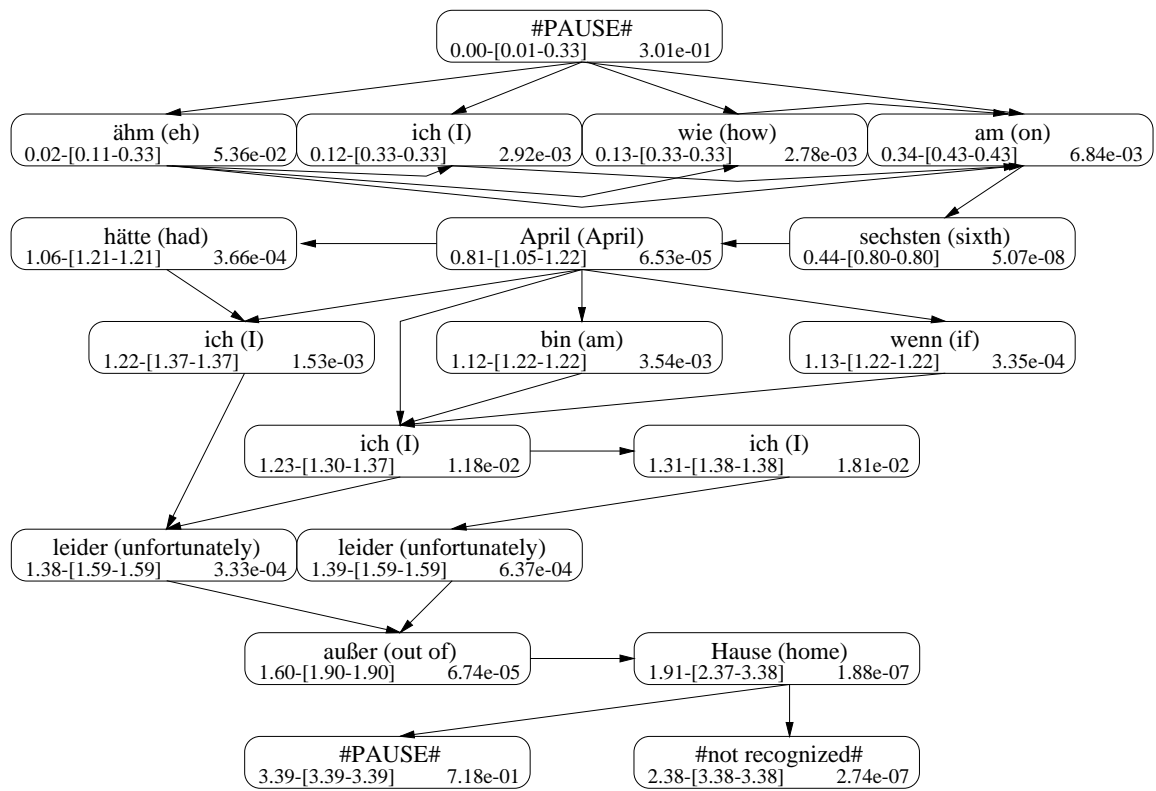

Figure 4: Simple word graph for a spoken utterance: "ähm am sechsten April bin ich leider außer Hause" ("eh on sixth April am I unfortunately out of home"). Each node represents a word hypothesis; each arrow represents its possible subsequent word hypotheses. Each word hypothesis is shown with its word string, start time, end time interval and acoustic plausibility.

These word hypotheses can overlap in time and constitute a directed graph called word graph. Each node in this word graph represents one word hypothesis. Two hypotheses in this graph of generated word hypotheses can be connected if the end time of the first word hypothesis is directly before the start time of the second word hypothesis. For instance, the word hypothesis for "am" ("on") ending at 0.43 and the hypothesis "sechsten" ("sixth") starting at 0.44 can be connected to a word hypothesis sequence.

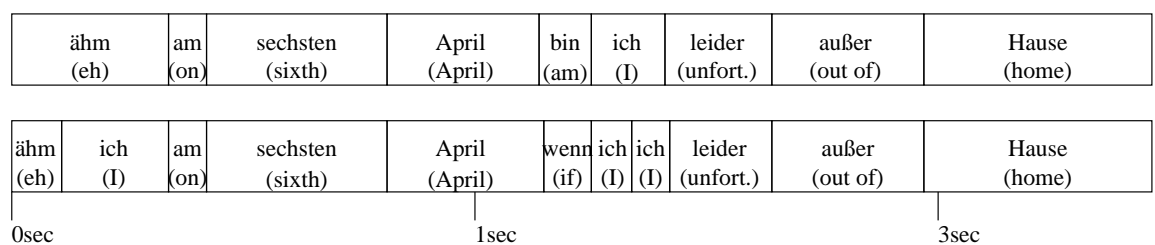

Figure 5: Two examples for word hypothesis sequences in a word graph

Our example word graph is very simple. However, as shown in Figure 5, a possible word hypothesis sequence is not only the desired "Ähm am sechsten April bin ich leider außer Hause" ("Eh on sixth April am I unfortunately out of home"), but also the sequence "Ähm ich am sechsten April wenn ich ich leider außer Hause" ("Eh I on sixth April if I I unfortunately out of home"). Consequently, we have to deal with incorrectly recognized words in an extraordinary order. Therefore syntactic and semantic analysis has to be very fault-tolerant in order to process such noisy word hypothesis sequences.

## 3. *Flat* Category Representation: An Intermediate Connecting Representation

In this section we will describe our flat category representations. First, we will show the categories for the syntactic analysis before we will depict the categories for the semantic analysis.

---

3. The speech input in the form of test word graphs was taken from the so-called Blaubeuren Meeting Corpus. The particular word graphs we used here were provided by project partners for general test purposes in the Verbmobil project. They were particularly generated for testing parsing strategies. Therefore the speech recognizer was fine-tuned to produce relatively small word graphs with a relatively high word accuracy of 93%. The vocabulary size for the HMM recognizer is 628. The average number of hypotheses per word was 6.3 over 10 dialogs.

## 3.1 Categories for Flat Syntactic Analysis

Flat syntactic analysis is the assignment of syntactic categories to a sequence of words, e.g., the word hypothesis sequence generated by a speech recognizer. Flat representations up to the phrase group level support local structural decisions. Local structural decisions deal with the problem of which phrase group (abstract syntactic category) a word belongs to. In this case the local, directly preceding words and their phrase group can influence the current decision. For instance, a determiner "the" could be part of a prepositional group "in the mine" or part of a starting noun group "the old mine". That is, local structural decisions depending on local context will be made based on a flat analysis.

For flat syntactic analysis we have developed a level of basic syntactic categories and abstract syntactic categories. These syntactic categories may vary depending on the language, and the degree of detail of the intended structural representation. However, the general approach is rather independent of the specifically used categories. In fact, we have used the same syntactic categories for two different domains: railway counter interactions and business meeting arrangements. The *basic syntactic categories* we used were noun, verb, preposition, pronoun, numeral, past participle, pause, adjective, adverb, conjunction, determiner, interjection and other. They are shown with their abbreviations in Table 1.

| Category | Examples | Category | Examples |
|---|---|---|---|
| noun (N) | date, April | adjective (J) | late |
| verb (V) | meet, choose | adverb (A) | often |
| preposition (R) | at, in | conjunction (C) | and, but |
| pronoun (U) | I, you | determiner (D) | the, a |
| numeral (M) | fourteenth | interjection (I) | eh, oh |
| participle (P) | taken | other (O) | particles |
| pause (/) | pause | | |

Table 1: Basic syntactic categories

The *abstract syntactic categories* we used are verb group, noun group, adverbial group, prepositional group, conjunction group, modus group, special group and interjection group. These abstract syntactic categories are shown in Table 2.

| Category | Examples |
|---|---|
| verb group (VG) | mean, would propose |
| noun group (NG) | a date, the next possible slot |
| adverbial group (AG) | later, as early as possible |
| prepositional group (PG) | in the dining hall |
| conjunction group (CG) | and, either ... or |
| modus group (MG) | interrogatives, confirmations: when, how long, yes |
| special group (SG) | additives like politeness: please, then |
| interjection group (IG) | interjections, pauses: eh, oh |

Table 2: Abstract syntactic categories

The categories should express main syntactic properties of the phrases. Most of our basic and abstract syntactic categories are widely used in different parsers. However, the approach of flat representations does not crucially rely on this specific set of basic and abstract syntactic categories. Our goal is to train, learn and generalize a flat syntactic analysis based on abstract syntactic categories and basic syntactic categories. Local syntactic decisions should be made as far as possible. Local syntactic ambiguities up to the phrase group level (abstract syntactic categories) can be dealt with but more global ambiguities like prepositional phrase attachment will not be dealt with since they will need additional knowledge, e.g., from a semantics module. While complete syntax trees have a certain preference (which might turn out to be wrong based on semantic knowledge), a flat syntactic representation goes as far as possible using only local syntactic knowledge for disambiguation.

## 3.2 Categories for Flat Semantic Analysis

Since semantic analysis is domain-dependent, the semantic categories can differ for different domains. We have worked particularly on two domains: railway counter interactions (called: Regensburg train corpus) and business meeting arrangements (called: Blaubeuren meeting corpus). There was about 3/4 overlap between the semantic categories of the train corpus and the meeting corpus (Wermter & Weber, 1996b). Differences occurred mainly for verbs, e.g., NEED-events are very frequent in the railway counter interactions while SUGGEST-events are frequent in the business meeting interactions. The semantic categories of the railway counter interactions were described in previous work (Weber & Wermter, 1995). Here we will primarily focus on the semantic categories of the meeting corpus. The *basic semantic categories* for a word are shown in Table 3. At a higher level of abstraction, each word can belong to an abstract semantic category. The possible abstract semantic categories are shown in Table 4. In summary, these categories provide a basis for a flat analysis. Each word is represented syntactically and semantically in its context by four categories at two basic and two abstract levels.

| Category | Examples |
|---|---|
| select (SEL) | select, choose |
| suggest (SUG) | propose, suggest |
| meet (MEET) | meet, join |
| utter (UTTER) | say, think |
| is (IS) | is, was |
| have (HAVE) | had, have |
| move (MOVE) | come, go |
| aux (AUX) | would, could |
| question (QUEST) | question words: where, when |
| physical (PHYS) | physical objects: building, office |
| animate (ANIM) | animate objects: I, you |
| abstract (ABS) | abstract objects: date |
| here (HERE) | time or location state words, prepositions: at, in |
| source (SRC) | time or location source words, prepositions: from |
| destination (DEST) | time or location destination words, prepositions: to |
| location (LOC) | Hamburg, Pittsburgh |
| time (TIME) | tomorrow, at 3 o' clock, April |
| negative evaluation (NO) | no, bad |
| positive evaluation (YES) | yes, good |
| nil (NIL) | words "without" specific semantics, e.g., determiner: a |

Table 3: Basic semantic categories

| Category | Examples |
|----------|----------|
| action (ACT) | action for full verb events: meet, select |
| aux-action (AUX) | auxiliary action for auxiliary events: would like |
| agent (AGENT) | agent of an action: I |
| object (OBJ) | object of an action: a date |
| recipient (RECIP) | recipient of an action: to me |
| instrument (INSTR) | instrument for an action: using an elevator |
| manner (MANNER) | how to achieve an action: without changing rooms |
| time-at (TM-AT) | at what time: in the morning |
| time-from (TM-FRM) | start time: after 6am |
| time-to (TM-TO) | end time: before 8pm |
| loc-at (LC-AT) | at which location: in Frankfurt, in New York |
| loc-from (LC-FRM) | start location: from Boston, from Dortmund |
| loc-to (LC-TO) | end location: to Hamburg |
| confirmation (CONF) | confirmation phrase: ok great, yes wonderful |
| negation (NEG) | negation phrase: no stop, not |
| question (QUEST) | question phrases: at what time |
| misc (MISC) | miscellaneous words, e.g., for politeness: please, eh |

Table 4: Abstract semantic categories

## 4. The Architecture of the SCREEN System

In this section we want to describe the constraints and principles which are important for our system design. As we outlined and motivated in the introduction, the *screening approach* is a flat, robust, learned analysis of spoken language based on category sequences (called flat representations) at various syntactic and semantic levels. In order to test this screening approach, we designed and implemented the hybrid connectionist SCREEN system which processes spontaneously spoken language by using learned connectionist flat representations. Here we summarize our main requirements in order to motivate the specific system design which will be explained in the subsequent subsections.

### 4.1 General Motivation for the Architecture

We consider *learning* to be extremely important for spoken-language analysis for several reasons. Learning reduces knowledge acquisition and increases portability, particularly in spoken-language analysis, where the underlying rules and regularities are difficult to formulate and often not reliable. Furthermore, in some cases, inductive learning may detect unknown implicit regularities. We want to use connectionist learning in simple recurrent networks rather than other forms of learning (e.g., decision trees) primarily because of the inherent fault-tolerance of connectionist networks, but also because knowledge about the sequence of words and categories can be learned in simple recurrent networks.

*Fault-tolerance* for often occurring language errors should be reflected in the system design. We do this for the commonly occurring errors (interjections, pauses, word repairs, phrase repairs). However, fault-tolerance cannot go so far as to try to model each class of occurring errors. The number of potentially occurring errors and unpredictable constructions is far too large. In SCREEN, we want to incorporate explicit fault-tolerance by using specific modules for correction as well as implicit fault-tolerance by using connectionist network techniques which are inherently fault-tolerant due to their support of similarity-based processing. In fact, even if a word is completely unknown, recurrent networks can use an empty input and may even assign the correct category if there is sufficient previous context.

*Flat representations*, as motivated in Sections 1 and 3, may support a robust spoken-language analysis. However, flat connectionist representations do not provide the full recursive power of arbitrary syntactic or semantic symbolic knowledge structures. In contrast to context-free parsers, flat representations provide a better basis for robust processing and automatic knowledge acquisition by inductive learning. However, it can also be argued that the use of potentially unrestricted recursion of well-known context-free grammar parsers provides a computational model with more recursive power than humans have in order to understand language. In order to better support robustness, we want to use flat representations for spontaneous language analysis.

*Incremental processing* of speech, syntax, semantics and dialog processing in parallel allows us to start the language analysis in parallel before the speech recognizer has finished its analysis. This incremental processing has the advantage of providing analysis results at a very early stage. For example, syntactic and semantic processing occur in parallel only slightly behind speech processing. When analyzing spoken language based on speech recognizer output, we want to consider many competing paths of word hypothesis sequences in parallel.

With respect to *hybrid representations*, we examine a hybrid connectionist architecture using connectionist networks where they are useful but we also want to use symbolic processing wherever necessary. Symbolic processing can be very useful for the complex control in a large system. On the other hand for learning robust analysis, we use feedforward and simple recurrent networks in many modules and try to use rather homogeneous, supervised networks.

## 4.2 An Overview of the Architecture

SCREEN has a parallel integrated hybrid architecture (Wermter, 1994) which has various main properties:

1. Outside of a module, there is no difference in communication between a symbolic and a connectionist module. While previous hybrid architectures emphasized different symbolic and connectionist representations, the different representations in SCREEN benefit from a common module interface. Outside of a connectionist or symbolic module all communication is identically realized by symbolic lists which contain values of connectionist units.

2. While previous hybrid symbolic and connectionist architectures are usually within either a symbolic or a connectionist module (Hendler, 1989; Faisal & Kwasny, 1990; Medsker, 1994), in SCREEN a global state is described as a collection of individual symbolic and connectionist modules. Processing can be parallel as long as one module does not need input from a second module.

3. The communication among the symbolic and connectionist modules is organized via messages. While other hybrid architectures have often used either only activation values or only symbolic structures, we used messages consisting of lists of symbols with associated activation or plausibility values to provide a communication medium which supports both connectionist processing as well as symbolic processing.

We will now give an overview of the various parts in SCREEN (see Figure 6). The important output consists of flat syntactic and semantic category representations based on the input of incrementally recognized parallel word hypotheses. A speech recognizer generates many incorrect word hypotheses over time, and even correctly recognized speech can contain many errors introduced by humans. A flat representation is used since it is more fault-tolerant and robust than, for instance, a context-free tree representation since a tree representation requires many more decisions than a flat representation.

Each module in the system, for instance the disambiguation of abstract syntactic categories, contains a connectionist network or a symbolic program. The integration of symbolic and connectionist representations occurs as an encapsulation of symbolic and connectionist processes at the module level. Connectionist networks are embedded in symbolic modules which can communicate with each other via messages.

However, what are the essential parts needed for our purposes of learning spoken-language analysis and why? Starting from the output of individual word hypotheses of a speech recognizer, we first need a component which receives an incremental stream of individual parallel word hypotheses and produces an incremental stream of word hypothesis sequences (see Figure 6). We call this part the *speech sequence construction part*. It is needed for transforming parallel overlapping individual word hypotheses to word hypothesis sequences. These word hypothesis sequences have a different quality and the goal is to find and work with the best word hypothesis sequences. Therefore we need a *speech evaluation part* which can combine speech-related plausibilities with syntactic and semantic plausibilities in order to restrict the attention to the best found word hypothesis sequences.

Furthermore, we need a part which analyzes the best found word hypothesis sequences according to their flat syntactic and semantic representation. The *category part* receives a stream of current word hypothesis sequences. Two such word hypothesis sequences are shown in Figure 6. This part provides the interpretation of a word hypothesis sequence with its basic syntactic categories, abstract syntactic categories, basic semantic categories, and abstract semantic categories. That is, each word hypothesis sequence is assigned four graded preferences for four word categories.

Human speech analyzed by a speech recognizer may contain many errors. So the question arises to what extent we want to consider these errors. An analysis of several hundred

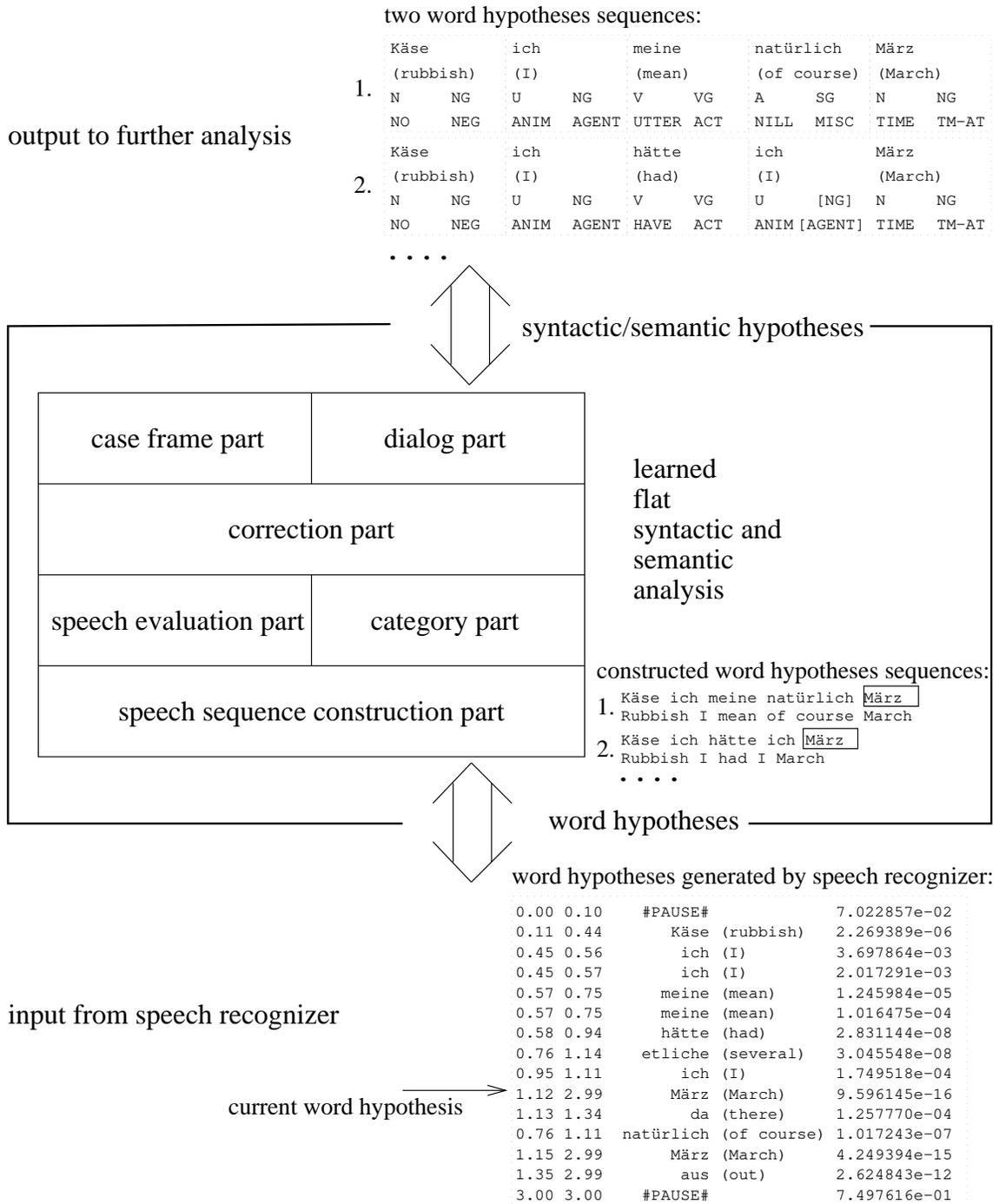

two word hypotheses sequences:

```
      Käse        ich        meine       natürlich    März
1.    (rubbish)   (I)        (mean)      (of course)  (March)
      N    NG     U    NG    V    VG     A    SG      N    NG
      NO   NEG    ANIM AGENT UTTER ACT   NILL MISC    TIME TM-AT
      Käse        ich        hätte       ich          März
2.    (rubbish)   (I)        (had)       (I)          (March)
      N    NG     U    NG    V    VG     U    [NG]     N    NG
      NO   NEG    ANIM AGENT HAVE ACT    ANIM [AGENT] TIME TM-AT
      . . . .
```

output to further analysis

syntactic/semantic hypotheses

| case frame part | dialog part |
|---|---|
| correction part | |
| speech evaluation part | category part |
| speech sequence construction part | |

learned
flat
syntactic and
semantic
analysis

constructed word hypotheses sequences:

```
1 Käse ich meine natürlich März
· Rubbish I mean of course March
2 Käse ich hätte ich März
· Rubbish I had I March
  . . . .
```

word hypotheses

word hypotheses generated by speech recognizer:

```
0.00 0.10    #PAUSE#                7.022857e-02
0.11 0.44       Käse (rubbish)      2.269389e-06
0.45 0.56        ich (I)            3.697864e-03
0.45 0.57        ich (I)            2.017291e-03
0.57 0.75      meine (mean)         1.245984e-05
0.57 0.75      meine (mean)         1.016475e-04
0.58 0.94      hätte (had)          2.831144e-08
0.76 1.14    etliche (several)      3.045548e-08
0.95 1.11        ich (I)            1.749518e-04
1.12 2.99       März (March)        9.596145e-16
1.13 1.34         da (there)        1.257770e-04
0.76 1.11  natürlich (of course)   1.017243e-07
1.15 2.99       März (March)        4.249394e-15
1.35 2.99        aus (out)          2.624843e-12
3.00 3.00    #PAUSE#                7.497616e-01
```

input from speech recognizer

current word hypothesis

Figure 6: Overview of SCREEN

transcripts and speech recognizer outputs revealed that there are some errors which occur often and regularly. These are interjections, pauses, word repairs, and phrase repairs. Therefore we designed a *correction part* which receives hypotheses about words and deals with most frequently occurring errors in spoken language explicitly.

These parts outlined so far build the center of the integration of speech-related and language-related knowledge in a flat fault-tolerant learning architecture, and therefore we will focus on these parts in this paper. However, if we want to process complete dialog turns which can contain several individual utterances we need to know where a certain utterance starts and which constituents belong to this utterance. This task is performed by a *case frame part* which fills a frame incrementally and segments a speaker's turn into utterances.

The long-term perspective of SCREEN is to provide an analysis for tasks such as spoken utterance translation or information extraction. Besides the syntactic and semantic analysis of an utterance, the intended dialog acts convey important additional knowledge. Therefore, a *dialog part* is needed for assigning dialog acts to utterances, for instance if an utterance is a request or suggestion. In fact, we have already fully implemented the case frame part and the dialog part for all our utterances. However, we will not describe the details of these two parts in this paper since they have been described elsewhere (Wermter & Löchel, 1996).

Learning in SCREEN is based on concepts of supervised learning as for instance in feed-forward networks (Rumelhart et al., 1986), simple recurrent networks (Elman, 1990) and more general recurrent plausibility networks (Wermter, 1995). In general, recurrent plausibility networks allow an arbitrary number of context and hidden layers for considering long distance dependencies. However, for the many network modules in SCREEN we attempted to keep the individual networks simple and homogeneous. Therefore, in our first version described here we used only variations of feedforward networks (Rumelhart et al., 1986) and simple recurrent networks (Elman, 1990). Due to their greater potential for sequential context representations, recurrent plausibility networks might provide improvements and optimizations of simple recurrent networks. However, for now we are primarily interested in an overall real-world hybrid connectionist architecture SCREEN rather than the optimization of single networks. In the following description we will give detailed examples of the individual networks.

## 4.3 A More Detailed View

After we motivated the various parts in SCREEN, we will now give a more detailed description of the architecture of SCREEN with respect to the modules for flat syntactic and semantic analysis of word hypothesis sequences. Therefore, we will focus on the speech related parts, the categorization part and correction part. Figure 7 shows a more detailed overview of these parts. The basic data flow is shown with arrows. Many modules generate hypotheses which are used in subsequent modules at a higher level. These hypotheses are illustrated with rising arrows. In some modules, the output contains local predictive hypotheses (sometimes called local top-down hypotheses) which are used again in modules at a lower level. These hypotheses are illustrated with falling arrows. Local predictive hypotheses are used in the correction part to eliminate[4] repaired utterance parts and in the speech evaluation part to eliminate syntactically or semantically implausible word hypothesis sequences. In some cases where arrows would have been too complex we have used numbers to illustrate the data flow between individual modules.

---

4. This means that repaired utterance parts are actually only marked as deleted.

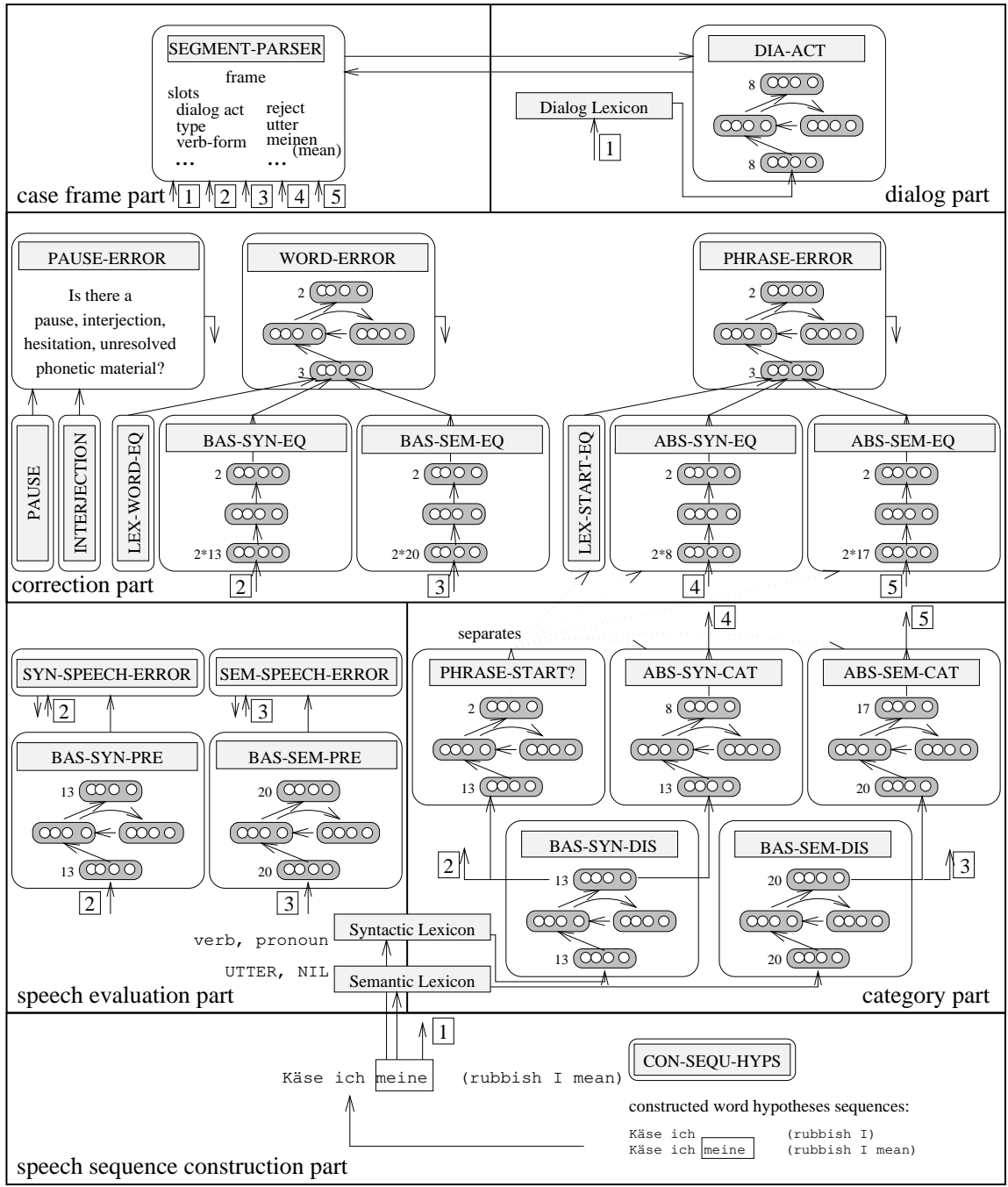

Figure 7: More detailed overview of SCREEN. The abbreviations and functionality of the modules are described in the text.

### 4.3.1 SPEECH SEQUENCE CONSTRUCTION PART

The speech sequence construction part receives a stream of parallel word hypotheses and generates a stream of word hypothesis sequences within the module CON-SEQU-HYPS at the bottom of Figure 7. Based on the current word hypotheses many word hypothesis sequences may be possible. In some cases we can reduce the number of current word hypotheses, e.g., if we know that time has passed so far that a specific word hypothesis sequence cannot be extended anymore at the time of the current word hypothesis. In this case we can eliminate this sequence since only word hypothesis sequences which could reach the end of the sentence are candidates for a successful speech interpretation.

Furthermore, we can use the speech plausibility values of the individual word hypothesis to determine the speech plausibility of a word hypothesis sequence. By using only some of the best word hypothesis sequences we can reduce the large space of possible sequences. The generated stream of word hypothesis sequences is similar to *a set of partial* N-best representations which are generated and pruned incrementally during speech analysis rather than at the end of the speech analysis process.

### 4.3.2 SPEECH EVALUATION PART

The speech evaluation part computes plausibilities based on syntactic and semantic knowledge in order to evaluate word hypothesis sequences. This part contains the modules for the detection of speech-related errors. Currently, the performance of speech recognizers for spontaneously spoken speaker-independent speech is in general still far from perfect. Typically, many word hypotheses are generated for a certain signal[5]. Therefore, many hypothesized words produced by a speech recognizer are incorrect and the speech confidence value for a word hypothesis alone does not provide enough evidence for finding the desired string for a signal. Therefore the goal of the speech evaluation part is to provide a preference for filtering out unlikely word hypothesis sequences. SYN-SPEECH-ERROR and SEM-SPEECH-ERROR are two modules which decide if the current word hypothesis is a syntactically (semantically) plausible extension of the current word hypothesis sequence. The syntactic (semantic) plausibility is based on a basic syntactic (semantic) category disambiguation and prediction.

In summary, each word hypothesis sequence has an acoustic confidence based on the speech recognizer, a syntactic confidence based on SYN-SPEECH-ERROR, and a semantic confidence based on SEM-SPEECH-ERROR. These three values are integrated and weighted equally[6] to determine the best word hypothesis sequences. That way, these two modules can act as an evaluator for the speech recognizer as well as a filter for the language processing part.

---

5. The HMM-speech recognizer used for generating word hypotheses in our domain has a word accuracy of about 93% for the best match between the word graph and the desired transcript utterance. This recognizer was particularly optimized for this task and domain in order to be able to examine the robustness at the language level. An unoptimized version for this task and domain currently has 72% word accuracy.

6. This integration of speech, syntax, and semantics confidence values provided better results than just using one or two of these three knowledge sources.

In statistical models for speech recognition, bigram or trigram models are used as language models for filtering out the best possible hypotheses. We used simple recurrent networks since these networks performed slightly better than a bigram and a trigram model which had been implemented for comparison (Sauerland, 1996). Later in Section 6.1 we will also show a detailed comparison of simple recurrent networks and n-gram models (for n = 1,...,5). The reason for this better performance is the internal representation of a simple recurrent network which does not restrict the covered context to a fixed number of two or three words but has the potential to learn the required context that is needed.

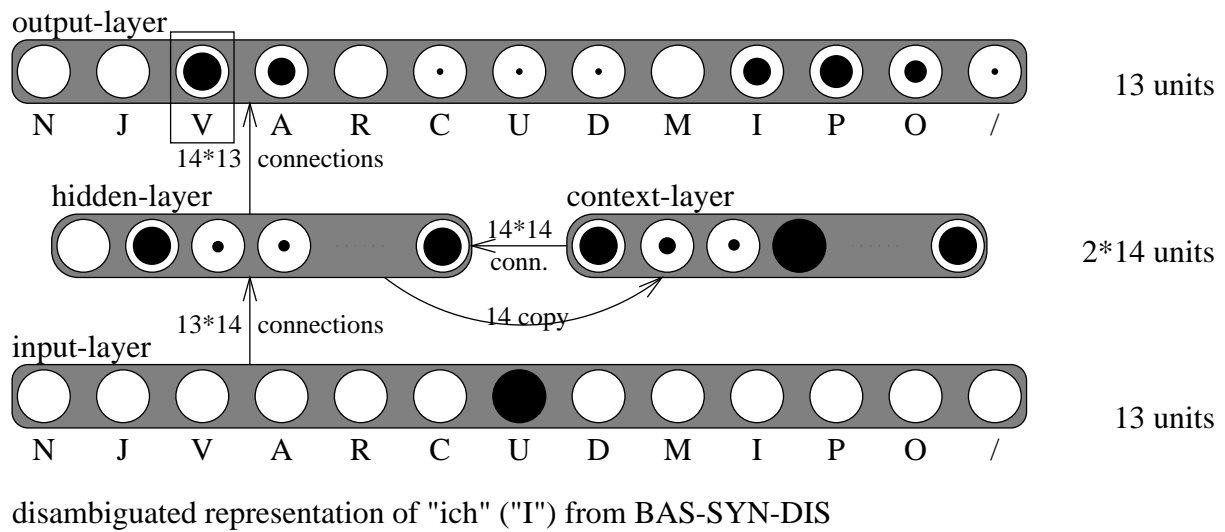

disambiguated representation of "ich" ("I") from BAS-SYN-DIS

Figure 8: Network architecture for the syntactic prediction in the speech evaluation part (BAS-SYN-PRE). The abbreviations are explained in Table 1.

The knowledge for the syntactic and semantic plausibility is provided by the prediction networks (BAS-SYN-PRE and BAS-SEM-PRE) of the speech evaluation part and the disambiguation networks (BAS-SYN-DIS and BAS-SEM-DIS) of the categorization part. As an example, we show the network for BAS-SYN-PRE in Figure 8. The previous basic syntactic category of the currently considered word hypothesis sequence is input to the network. In our example "ich" ("I") from the word hypothesis sequence "Käse ich meine" ("Rubbish I mean") is found to be a pronoun (U). Therefore, the syntactic category representation for "ich" ("I") contains a "1" for the pronoun (U) category. All other categories receive a "0".

The input to this network consists of 13 units for our 13 categories. The output of the network has the same size. Each unit of the vector represents a plausibility for the predicted basic syntax category of the last word in the current word hypothesis sequence. The plausibility of the unit representing the desired basic syntactic category (found by BAS-SYN-DIS) is taken as syntactic plausibility for the currently considered word hypothesis sequence by SYN-SPEECH-ERROR. In this example "meine" ("mean") is found to be a verb (V). Therefore the plausibility for a verb (V) will be taken as syntax plausibility (selection marked by a box in the output-layer of BAS-SYN-PRE in Figure 8).

In summary, the syntactic (semantic) plausibility of a word hypothesis sequence is evaluated by the degree of agreement between the disambiguated syntactic (semantic) category of the current word and the predicted syntactic (semantic) category of the previous word. Since decisions about the current state of a whole sequence have to be made, the preceding context is represented by copying the hidden layer for the current word to the context layer for the next word based on an SRN network structure (Elman, 1990). All connections in the network are n:m connections except for the connections between the hidden layer and the context layer which are simply used to copy and store the internal preceding state in the context layer for later processing when the next word comes in. In general, the speech evaluation part provides a ranking of the current word hypothesis sequences by the equally weighted combination of acoustic, syntactic, and semantic plausibility.

### 4.3.3 CATEGORY PART

The module BAS-SYN-DIS performs a basic syntactic disambiguation (see Figure 9). Input to this module is a sequence of potentially ambiguous syntactic word representations, one for each word of an utterance at a time. Then this module disambiguates the syntactic category representation according to the syntactic possibilities and the previous context. The output is a preference for a disambiguated syntactic category. This syntactic disambiguation task is learned in a simple recurrent network. Input and output of the network are the ambiguous and disambiguated syntactic category representations. In Figure 9 we show an example input representation for "meine" ("mean", "my") which can be a verb and a pronoun. However, in the sequence "Ich meine" ("I mean"), "meine" can only be a verb and therefore the network receives the disambiguated verb category representation alone.

The module BAS-SEM-DIS is similar to the module BAS-SYN-DIS but instead of receiving a potentially ambiguous syntactic category input and producing a disambiguated syntactic category output, the module BAS-SEM-DIS receives a semantic category representation from the lexicon and provides a disambiguated semantic category representation output. This semantic disambiguation is learned in a simple recurrent network which provides the mapping from the ambiguous semantic word representation to the disambiguated semantic word representation. Both modules BAS-SYN-DIS and BAS-SEM-DIS provide this disambiguation so that subsequent tasks like the association of abstract categories and the test of category equality for word error detection is possible.

The module ABS-SYN-CAT supplies the mapping from disambiguated basic syntactic category representations to the abstract syntactic category representations (see Figure 10). This module provides the abstract syntactic categorization and it is realized with a simple recurrent network. This module is important for providing a flat abstract interpretation of an utterance and for preparing input for the detection of phrase errors. Figure 10 shows that the disambiguated basic syntactic representation of "meine" ("mean") as a verb - and a very small preference for a pronoun - is mapped to the verb group category at the higher abstract syntactic category representation. Based on the number of our basic and abstract syntactic categories there are 13 input units for the basic syntactic categories and 8 output units for the abstract syntactic categories.

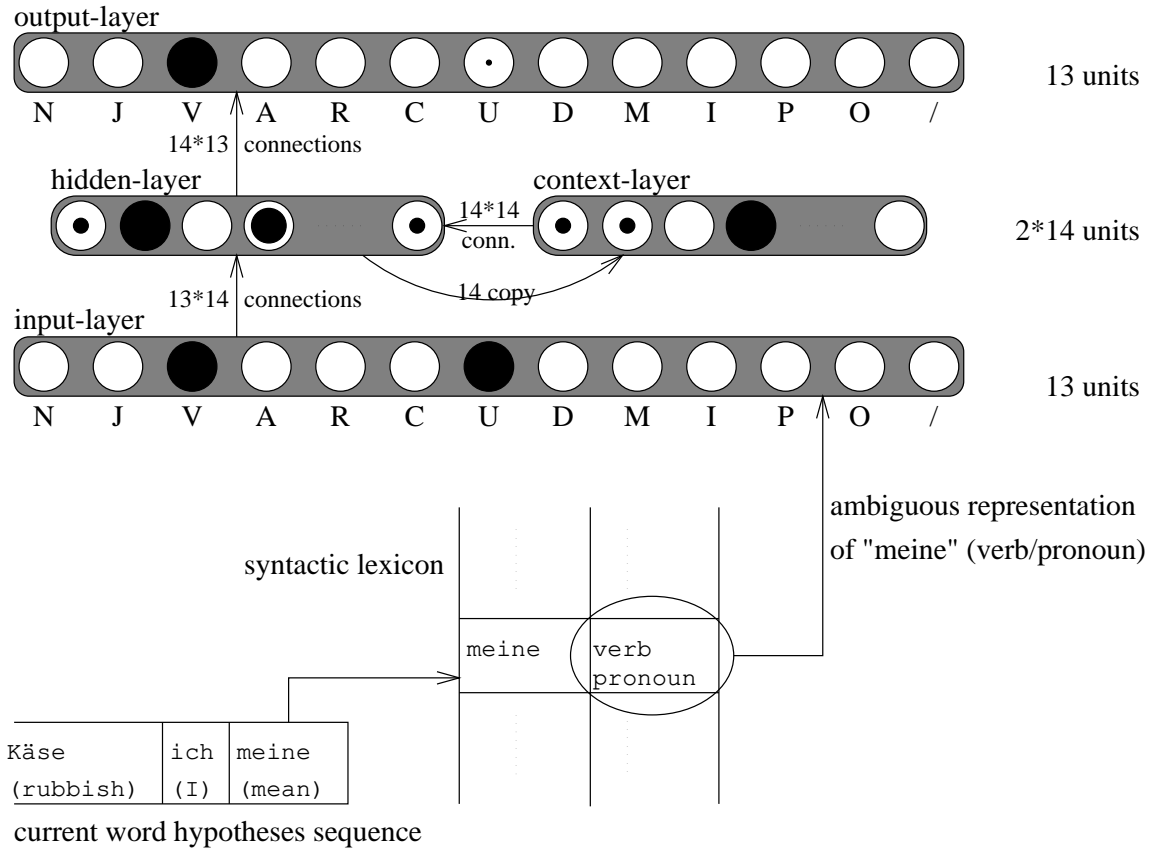

Figure 9: Network architecture for the basic syntactic disambiguation (BAS-SYN-DIS). The abbreviations are explained in Table 1.

The module ABS-SEM-CAT is a parallel module to ABS-SYN-CAT but uses basic semantic category representations as input and abstract semantic category representations as output. Similar to the previous modules, we also used a simple recurrent network to learn this mapping and to represent the sequential context. The input to the network is the basic semantic category representation for the word, and the output is an abstract category preference.

These described four networks provide the basis for the fault-tolerant flat analysis and the detection of errors. Furthermore, there is the module PHRASE-START for distinguishing abstract categories. The task of this module is to indicate the boundaries of subsequent abstract categories with a delimiter. We use these boundaries to determine the abstract syntactic and abstract semantic category of a phrase[7]. Earlier experiments had provided support to take the abstract syntactic category of the first word in a phrase as the final abstract syntactic category of a phrase, since phrase starts (e.g., prepositions) are good indicators for abstract syntactic categories (Wermter & Löchel, 1994). On the other hand, earlier experiments supported to take the abstract semantic category of the last word of a phrase as the final abstract semantic category of a phrase, since phrase ends (e.g., nouns) are good indicators for abstract semantic categories (Wermter & Peters, 1994). Furthermore, the phrase start gives us an opportunity to distinguish two equal subsequent abstract categories of two phrases. For instance, if we have a phrase like "in Hamburg on Monday" we have to know where the border exists between the first and the second prepositional phrase.

---

7. In Figure 7 we show the influence of the phrase start delimiter on the abstract syntactic and semantic categorization with dotted lines.

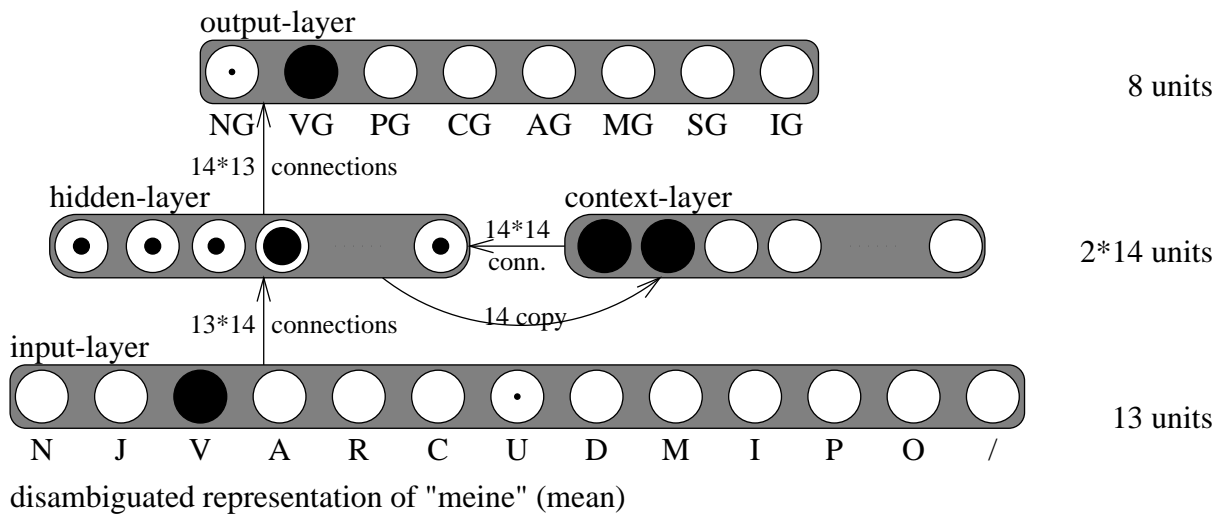

Figure 10: Network architecture for the abstract syntactic categorization (ABS-SYN-CAT). The abbreviations are explained in Table 2.

### 4.3.4 CORRECTION PART

The correction part contains modules for detecting pauses, interjections, as well as repetitions and repairs of words and phrases (see Figure 7). The modules for detecting pause errors are PAUSE-ERROR, PAUSE and INTERJECTION. The modules PAUSE and INTERJECTION receive the currently processed word and detect the potential occurrence of a pause and interjection, respectively. The output of these modules is input for the module PAUSE-ERROR. As soon as a pause or interjection has been detected, the word is marked as deleted and therefore virtually eliminated from the input stream[8]. An elimination of interjections and pauses is desired - for instance in a speech translation task - in order to provide an inter-pretation with as few errors as possible. Since these three modules are basically occurrence tests they have been realized with symbolic representations.

---

8. Pauses and interjections can sometimes provide clues for repairs (Nakatani & Hirschberg, 1993) although currently we do not use these clues for repair detection. Compared to the lexical, syntactic, and semantic equality of constituents, interjections and pauses provide relatively weak indicators for repairs since they also occur relatively often at other places in a sentence. However, since we just mark interjections and pauses as deleted we could make use of this knowledge in the future if necessary.

The second main cluster of modules in the correction part are the modules which are responsible for the detection of word-related errors. Then, word repairs as in "Am sechsten April bin <u>ich ich</u>" ("on sixth April am <u>I I</u>") or "Wir haben ein <u>Termin Treffen</u>" ("We have a <u>date meeting</u>") can be dealt with. There are certain preferences for finding repetitions and repairs at the word level. Among these preferences there is the lexical equality of two subsequent words (symbolic module LEX-WORD-EQ), the equality of two basic syntactic category representations (connectionist module BAS-SYN-EQ), and the equality of the basic semantic categories of two words (connectionist module BAS-SEM-EQ). As an example for the three modules, we show the test for syntactic equality (BAS-SYN-EQ) in Figure 11.

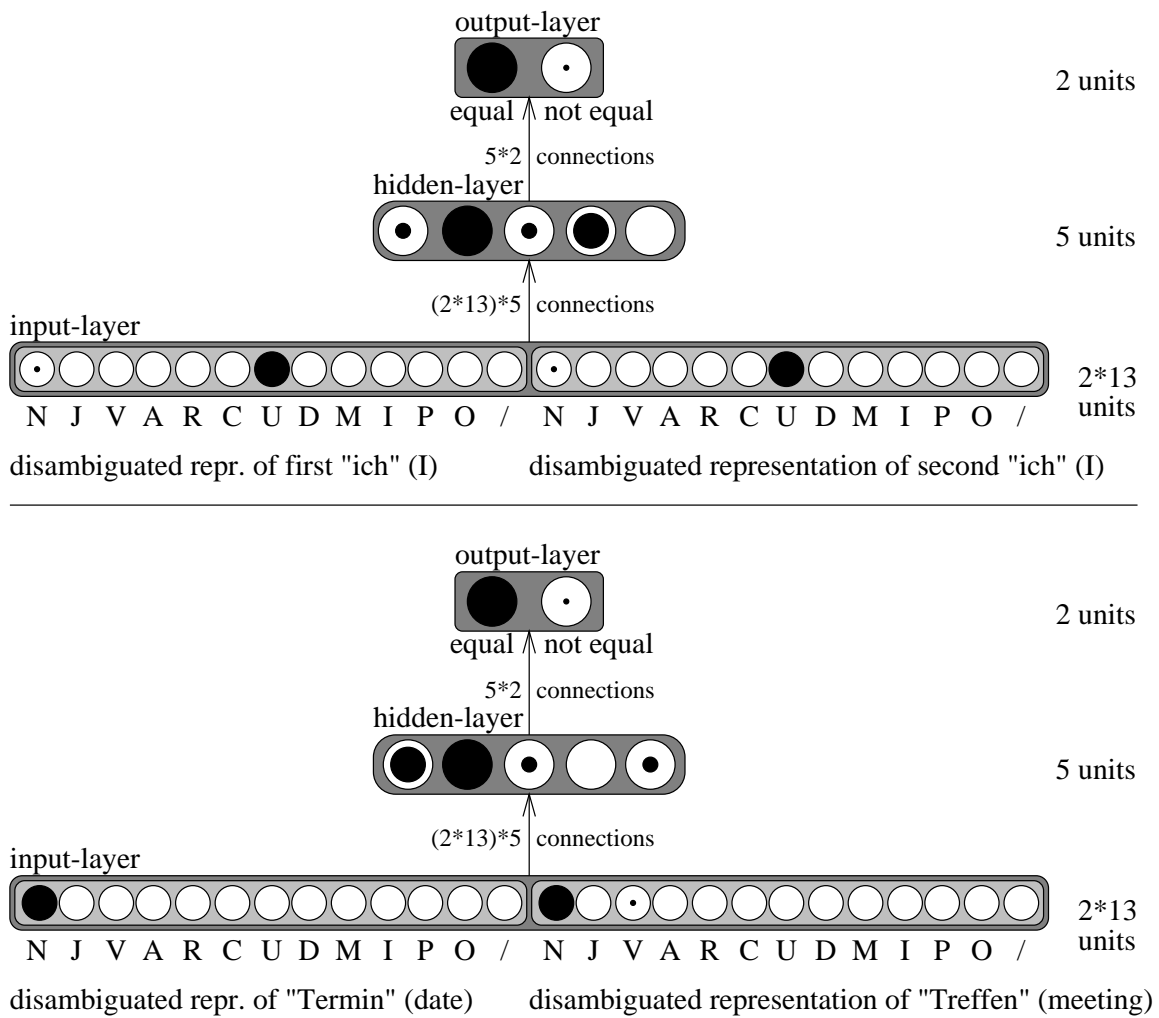

Figure 11: Network architecture for the equality of basic syntactic category representation (BAS-SYN-EQ). The abbreviations are explained in Table 1.

Two output units for plausible/implausible outcome have been used here since the network with two output units gave consistently better results compared with a network with only one output unit (with 1 for plausible and 0 for implausible). The reason why the network with two output units performed better is the separation of the weights for plausible and implausible in the hidden-output layer. In order to receive a single value, the two output values are integrated according to the formula: $unit_1 * (1.0 - unit_2)$. Then, the output of all three equality modules is a value between 0 and 1 where 1 represents equality and 0 represents inequality. Although a single such preference may not be sufficient, the common influence provides a reasonable basis for detecting word repairs and word repetitions in the module WORD-ERROR. Then, word repairs and repetitions are eliminated from the original utterance. Since the modules for word-related errors are based on two representations of two subsequent input words and since context can only play a minor role, we use feedforward networks for these modules. On the other hand, the simple test on lexical equality of the two words in LEX-WORD-EQ is represented more effectively using symbolic representation.

The third main cluster in the correction part consists of modules for the detection and correction of phrase errors. An example for a phrase error is: "Wir brauchen den früheren Termin den späteren Termin" ("We need the earlier date the later date"). There are preferences for phrase errors if the lexical start of two subsequent phrases is equal, if the abstract syntactic categories are equal and if the abstract semantic categories are equal. For these three preferences we have the modules LEX-START-EQ, ABS-SYN-EQ and ABS-SEM-EQ. All these modules receive two input representations of two corresponding words from two phrases, LEX-START-EQ receives two lexical words, ABS-SYN-EQ two abstract syntactic category representations, and ABS-SEM-EQ two abstract semantic category representations. The output of these three modules is a value toward 1 for equality and toward 0 otherwise. These values are input to the module PHRASE-ERROR which finally decides whether a phrase is replaced by another phrase. As the lexical equality of two words is a discrete test, we have implemented LEX-START-EQ symbolically, while the other preferences for a phrase error have been implemented as feedforward networks.

## 5. Detailed Analysis with Examples

In this section we will have a detailed look at processing the output from a speech recognizer and producing a flat syntactic and semantic interpretation of concurrent word hypothesis sequences (also called sentence hypothesis here).

### 5.1 The Overall Environment

The overall processing is incremental from left to right, and any time multiple sentence hypotheses are processed in parallel. Figure 12 shows a snapshot of SCREEN after 0.95s of the utterance. At this time the snapshot shows the first three sentence hypotheses as the German words together with their (literal) English translations ("Rubbish I mean", "Rubbish I", "Rubbish I had"). The SCREEN environment allows the user to view and inspect the incremental generation of word hypothesis sequences (partial sentence hypotheses) and their most preferred syntactic and semantic categories at the basic and abstract level. Each sentence hypothesis is illustrated horizontally. At a certain time many sentence hypotheses can be active in parallel. They are ranked according to the descending plausibility of the sentence hypotheses. So in the snapshot in Figure 12 there are currently three sentence hypotheses and the preferred current sentence hypothesis consists of "Rubbish I mean".

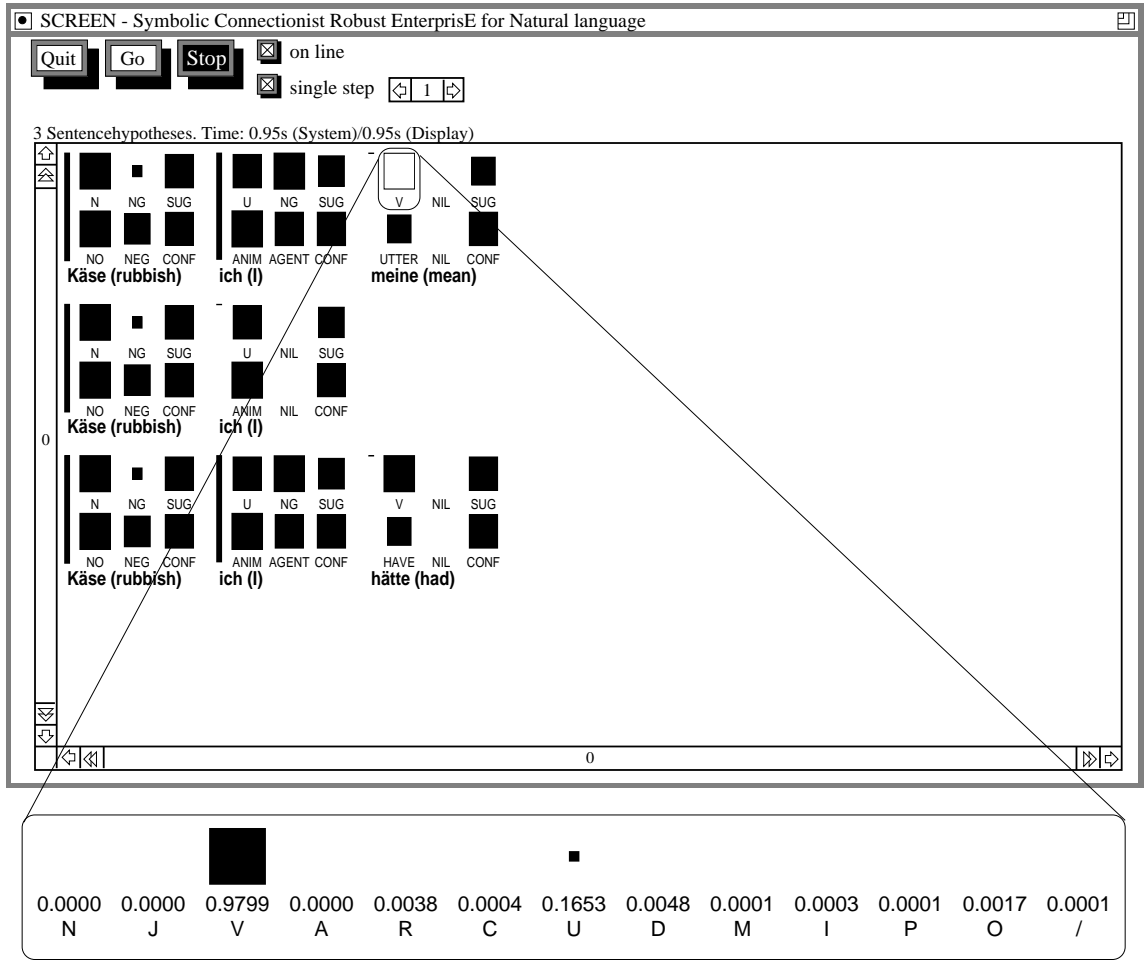

Figure 12: First snapshot for sentence "Käse ich meine natürlich März ("Rubbish I mean of course March"). The abbreviations are explained in Table 1 to 4. Below, the second pop-up window illustrates the full preferences of the word "meine" ("mean") for its basic syntactic categories.

All these sentence hypotheses are syntactically and semantically plausible starts. The underlying variations are introduced by the speech recognizer which produced different word hypotheses for slightly overlapping signal parts of the sentence. Besides the speech plausibility, syntax and also semantics can help with choosing better sentence hypotheses. Currently we combine the speech recognition plausibility, the syntactic plausibility, and the semantic plausibility to compute the plausibility of the sentence hypotheses as a multiplication of the respective normalized plausibility values between 0 and 1. Since the speech recognizer does not contain syntactic and semantic knowledge, a sequence hypothesis rated plausible based on speech knowledge alone may neglect the potential of syntactic and semantic regularity.

By using corresponding syntactic and semantic plausibility values for a sentence hypothesis we can integrate acoustic, syntactic, and semantic knowledge.

Each word hypothesis is shown with the preferred basic syntactic hypothesis (upper left square of a word hypothesis), the preferred abstract syntactic hypothesis (upper middle square), the preferred basic semantic hypothesis (lower left square), the preferred abstract semantic hypothesis (lower middle square), the preferred dialog act (upper right square)[9], and the integrated acoustic, syntactic and semantic confidence of the partial sentence hypothesis up to that point (lower right square). The size of the square illustrates the strength of the hypothesis, and a full black square means that a preferred hypothesis is close to one. For instance, in the word hypothesis for "ich" ("I") in the first sentence hypothesis we have the hypothesis of a pronoun (U) as the basic syntactic category, a noun group (NG) as the abstract syntactic category, an animate object (ANIM) as the basic semantic category, an AGENT as the abstract semantic category, and suggestion (SUG) as dialog act. Furthermore, the length of a vertical bar between word hypotheses indicate the plausibility for a new phrase start.

As another example, we can see the representation of our example word "meine" (could be the verb "mean" or the pronoun "my" in German) which we have used throughout the network descriptions (see Figure 9). The network had a correct preference for "meine" being a verb (V). Figure 12 shows this preference as well as a *zoomed* illustration of all other less favored preferences in a second pop-up window below. As we can see, the ambiguous other pronoun preference U received the second strongest activation while all other preferences are close to 0. These shown activation preferences are the output values of the corresponding network for basic syntactic categorization. So any shown activation value in our snapshots shows only the most preferred hypothesis while all other hypotheses can be shown on request[10].

Within the display we can scroll up and down the descending and ascending sentence hypotheses. Furthermore we can scroll left and right for analyzing specific longer word hypothesis sequences. There is also a step mode which allows the SCREEN system to wait for an interactive mouse click to process the next incoming word hypothesis for a very detailed analysis. This step mode can be adapted for a different number of steps (word hypotheses) and it can be switched off completely if one decides to analyze the sentence hypotheses later or at the end of all word hypotheses. Only the preferred of all possible syntactic and semantic hypotheses are shown. Therefore many different hypotheses appear to have the same size. However, by clicking on one of the squares the other less confident hypotheses can be displayed as well.

---

9. The dialog acts we use are: accept (ACC), query (QUERY), reject (REJ), request-suggest (RE-S), request-state (RE-S), state (STATE), suggest (SUG), and miscellaneous (MISC). Since this paper focuses on the syntactic and semantic aspects of SCREEN we do not further elaborate on the implemented dialog part here. Further details on dialog act processing have been described previously (Wermter & Löchel, 1996).

10. In the snapshots in Figure 12 the abstract syntactic and semantic categories have not yet been computed and therefore are represented as NIL. In the next processing step this computation will be performed which can be seen in next Figure 13.

## 5.2 Analyzing the Final Snapshot in Short Sentence Hypotheses

In Figure 13 we illustrate the final state after 3.01s of the utterance. Eight possible sentence hypotheses remained out of which we see the first four in Figure 13. Starting with the fourth sentence hypothesis "Käse ich hätte ich März" ("Rubbish I had I march") we can see that this lower rated sentence hypothesis is not the desired sentence. The lower ranked hypotheses are good examples that current state-of-the-art speech recognizers alone will not be able to produce reliable sentence hypotheses, since the problem of analyzing spontaneous speaker-independent speech is very complex. Therefore the syntactic and semantic components for spontaneous language have to take into account that there will be highly irregular sequences as shown below. However, it is interesting to observe that the underlying connectionist networks always produce a preference for the syntactic and semantic interpretation at the abstract and basic level. In fact, although the lower ranked sentence hypotheses do not constitute the desired sentence all assigned syntactic and semantic categories are correct for the individual word hypotheses. Of course there may be cases that a network also could make a wrong decision for uncertain word hypotheses. However the syntactic and semantic processing will never break for any possible sentence hypothesis, and is in this respect different from more well-known methods like symbolic context-free chart parsers.

If we look at the top-ranked sentence hypothesis "Käse ich meine natürlich März" ("Rubbish I mean of course March") this is also the desired sentence. It is the most plausible sentence based on speech and language plausibility. Furthermore, we can see that the assigned categories are correct: The German word "Käse" ("Rubbish") is found to be a noun as part of a noun group which expresses a negation. "Ich" ("I") starts a new phrase, that is a pronoun as a noun group which represents an animate being and an agent. The following German word "meine" is particularly interesting since it can be used as a verb in the sense of "mean" but also as a pronoun in the sense of "my". Therefore, the connectionist network for the basic syntactic classification has to disambiguate these two possibilities based on the preceding context. The network has learned to take into consideration the preceding context and is able to choose the correct basic syntactic category verb (V) rather than pronoun (U) for the word "meine" ("mean"). At this time a new phrase start has been found as well. The following word "natürlich" ("of course") has the highest preference for an adverb and a special group. Finally, the word "März" ("March") is assigned the highest plausibility for a noun and noun group as well as a time at which something happens.

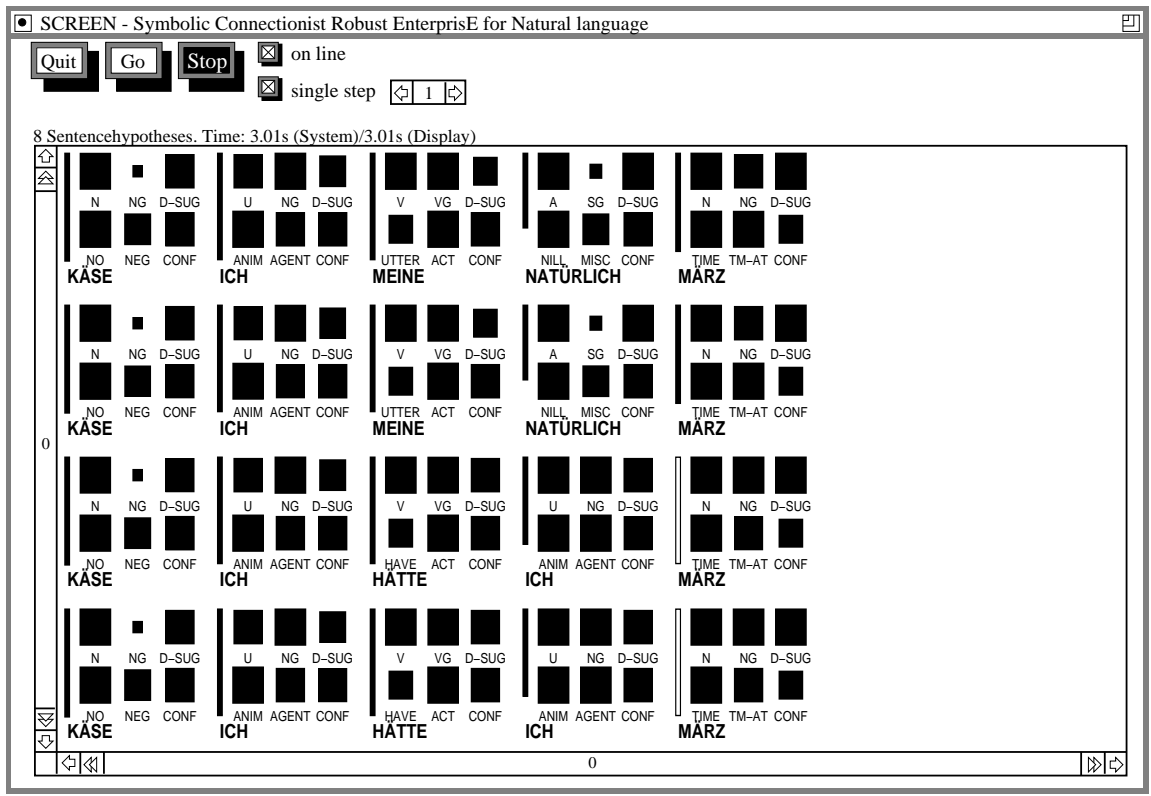

Figure 13: Final snapshot for sentence "Käse ich meine natürlich März ("Rubbish I mean of course March").

## 5.3 Phrase Starts and Phrase Groups in Longer Sentence Hypotheses

Now we will focus on a detailed analysis of a second example: "Ähm ja genau allerdings habe ich da von neun bis vier Uhr schon einen Arzttermin". The literally translated

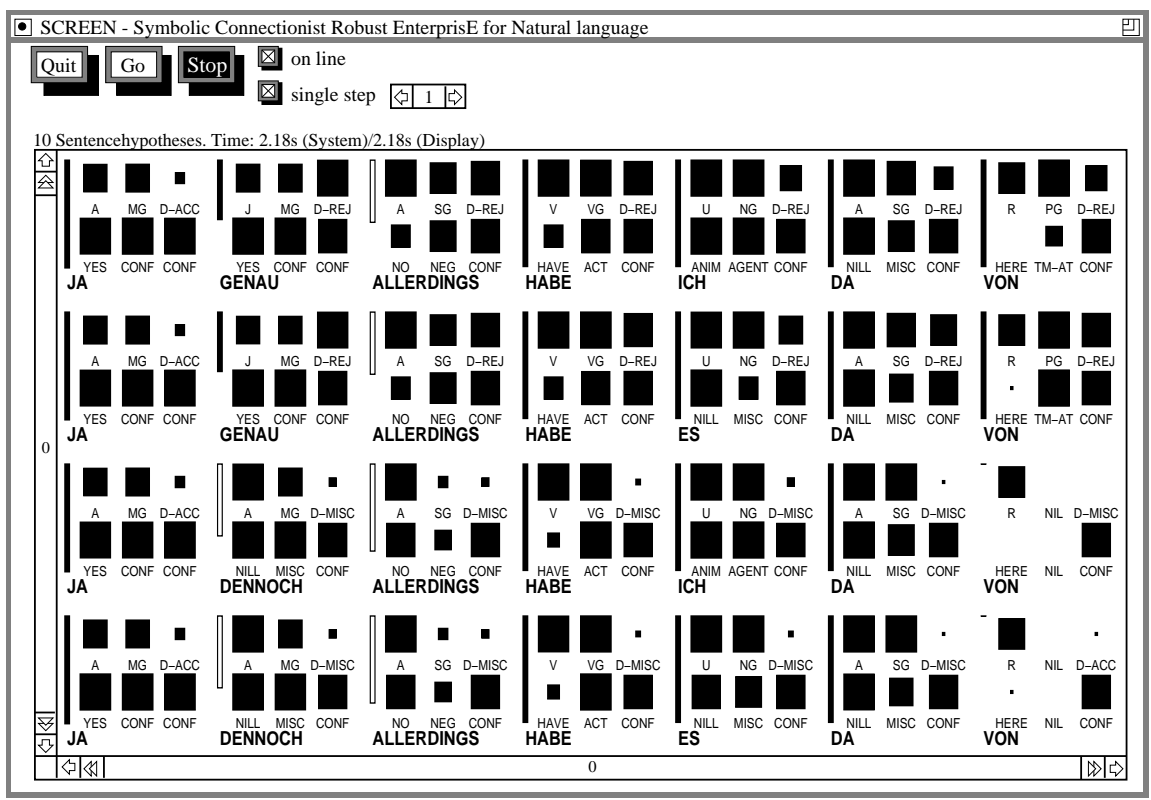

Figure 14: First part of the snapshot for sentence "Ähm ja genau allerdings habe ich da von neun bis vier Uhr schon einen Arzttermin" (literal translation: "Yes exactly however have I there from nine to four o'clock already a doctor-appointment"; improved translation: "Eh yes exactly however then I have a doctor appointment from nine to four o'clock").

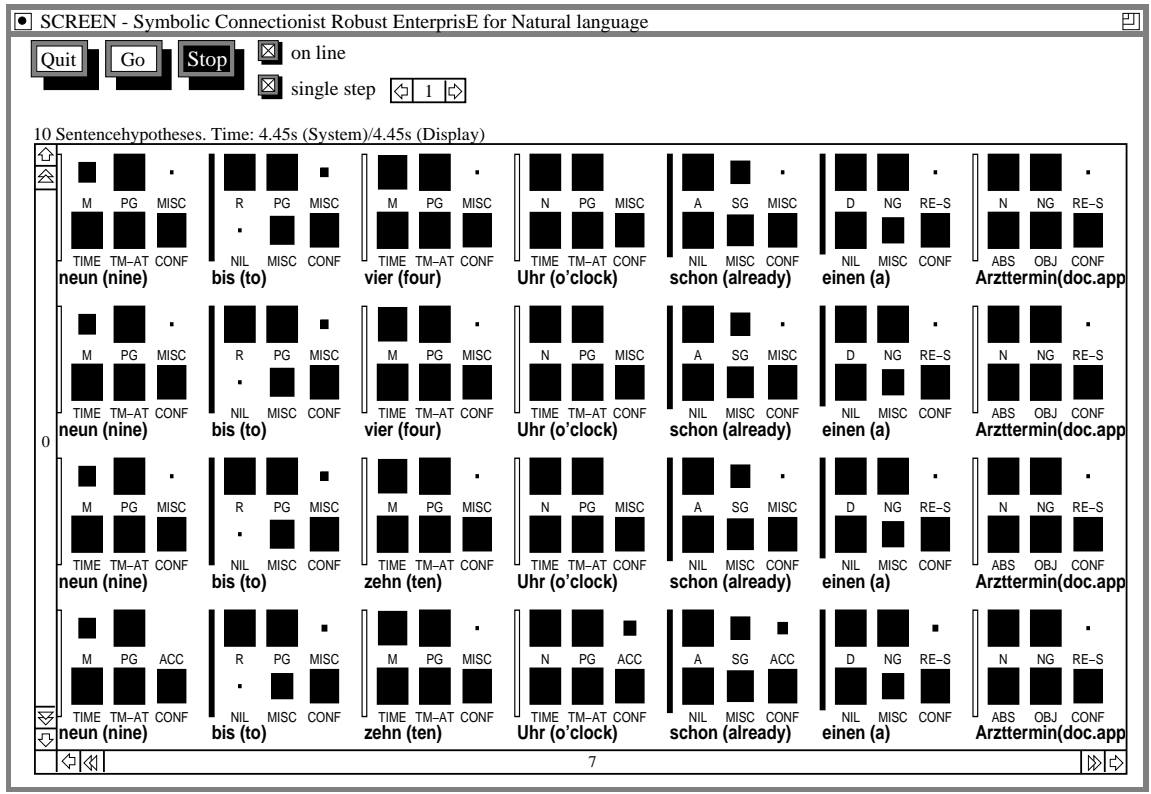

Figure 15: Second part of the snapshot for sentence "Ähm ja genau allerdings habe ich
da von neun bis vier Uhr schon einen Arzttermin" ("Yes exactly however have
I there from nine to four o'clock already a doctor-appointment").

sentence to be analyzed is: "Eh yes exactly however have I there from nine to four o'clock
already a doctor-appointment". A better but non-literal translation would be: "Eh yes
exactly however then I have a doctor appointment from nine to four o'clock". During
the analysis of the first few sentence hypotheses, the interjection "ähm" ("eh") is detected
by the corresponding module in the correction part and is eliminated from the respective
sentence hypotheses.

In Figure 14 and Figure 15 we show the best found four sentence hypotheses. The
categories of these sentence hypotheses look similar but we have to keep these separate
hypotheses since they differ in their time stamps and their speech confidence values.

In these two snapshots of this longer example we can also illustrate the influence of
the phrase starts. The sequences "von neun" ("from nine") and "bis vier Uhr" ("to four
o'clock") constitute two phrase groups which are clearly separated by the black bar before
the prepositions "von" ("from") and "bis" ("to"). All the other words "neun" ("nine"),
"vier" ("four"), and "Uhr" ("o'clock") do not start another phrase group. Since the under-
lying connectionist network for learning the phrase boundaries is a simple recurrent network
this example demonstrates that this network has learned the preceding context. Without
having learned that there had been a preposition "von" ("from") or "bis" ("to") a noun like "Uhr" ("o'clock") does not have to be within a prepositional phrase group but could also be part of a noun phrase in another context like "vier Uhr paßt gut" ("four o'clock fits well").

## 5.4 Dealing with Noise as Repairs

Finally we will focus on the example for the simple word graph shown in the beginning of this paper on page 41: "Ähm am sechsten April bin ich leider außer Hause". The literal translation is "Eh on 6th April am I unfortunately out of home". Using this sentence we will give an example for an interjection and a simple word repair. Dealing with hesitations and repairs is a large area in spontaneous language processing and is not the main topic of this paper (a more detailed discussion on repairs in SCREEN can be found in previous work, Weber & Wermter, 1996). Nevertheless, for the sake of illustration and completeness we show the ability of SCREEN to deal with interjections and word repairs. The first snapshot in Figure 16 shows the start of our example sentence after 1.39s. The leading interjection "eh" has been eliminated already.

Furthermore, we can see that the second word hypothesis sequence shows two subsequent word hypotheses for "ich" ("I"). This is possible since there were two word hypotheses

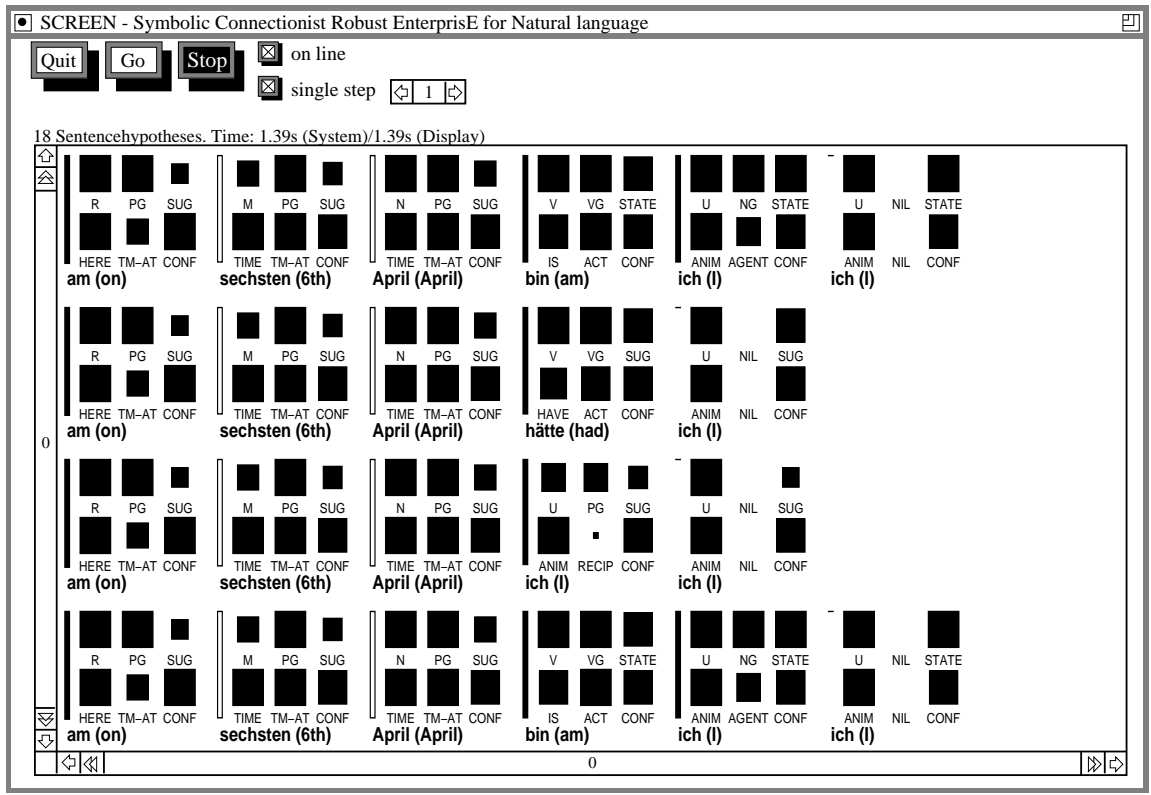

Figure 16: First snapshot for sentence "Ähm am sechsten April bin ich leider außer Hause" ("Eh on 6th April am I unfortunately out of home").

generated by the speech recognizer which could be connected. In this case there were the four word hypotheses shown below:

| start time | end time | word hypothesis | speech plausibility |
|---|---|---|---|
| 1.22s | 1.37s | ich (I) | 1.527688e-03 |
| 1.23s | 1.30s | ich (I) | 1.178415e-02 |
| 1.23s | 1.37s | ich (I) | 2.463924e-03 |
| 1.31s | 1.38s | ich (I) | 1.813340e-02 |

Just using this speech knowledge from the word hypotheses, it is possible to connect the second hypothesis which runs from 1.23s to 1.30s with the fourth hypothesis which runs from 1.31s to 1.38s. This is an example of noise generated by the speech recognizer, since the desired sentence contains only one word "ich" ("I") but the sentence hypothesis at this point contains two. This repetition can be treated and eliminated in the same way as actual word repairs in language. While the reasons for the occurrence of such repairs are different the effect of a repeated word is the same. Therefore, in this case the repeated "ich" ("I") is eliminated from the sentence sequence. In Figure 17 we show the final snapshot of the sentence. We can see that no word repairs occur in the top-ranked sentence hypothesis which is also the desired sentence.

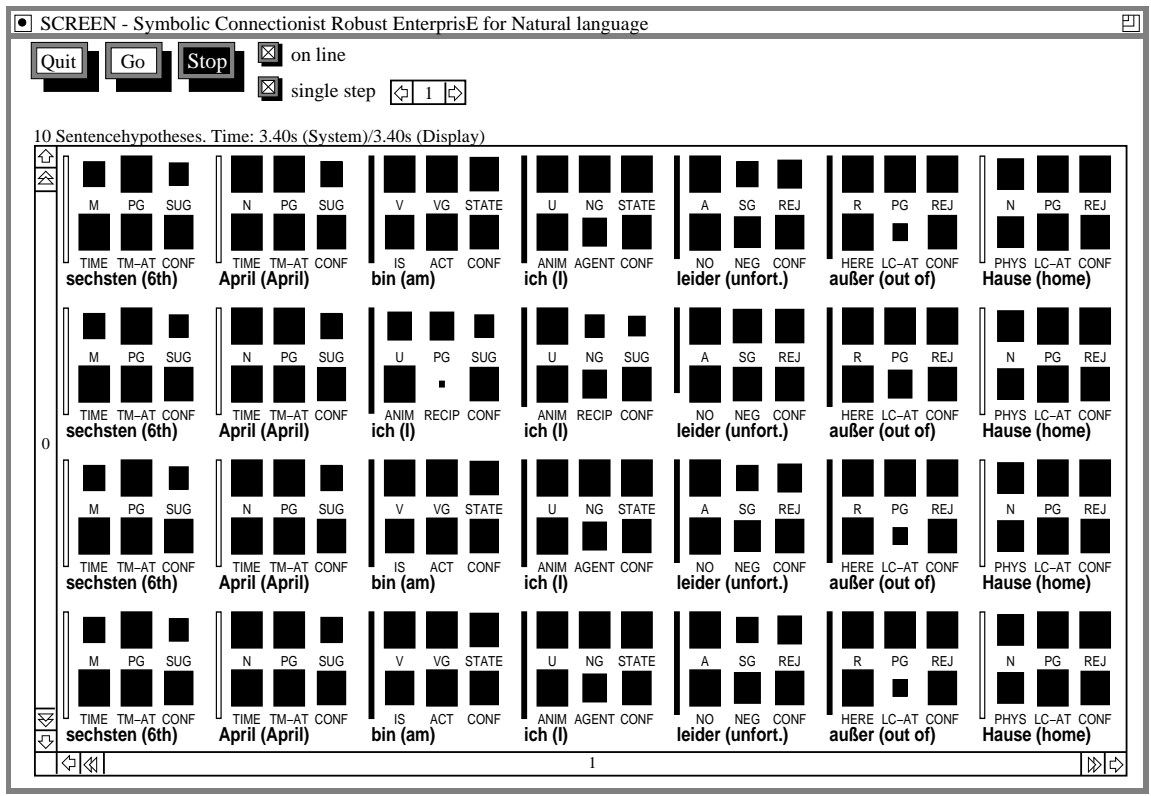

Figure 17: Final snapshot for sentence "Ähm am sechsten April bin ich leider außer Hause" ("Eh on 6th April am I unfortunately out of home").

In general, for language repairs, SCREEN can deal with the elimination of interjections and pauses, the repair of word repetitions, word corrections (where the words may be different, but their categories are the same) as well as simple forms of phrase repairs (where a phrase is repeated or replaced by another phrase).

## 6. Design Analysis of SCREEN

In this section we will describe our design choices in SCREEN. In particular we focus on the issues why we use connectionist networks, why we reach high accuracy with little training, and how SCREEN can be compared to other systems and other design principles.

### 6.1 Why Did We Use Connectionist Networks in SCREEN?

In the past, n-gram based techniques have been used successfully for tasks like syntactic category prediction or part of speech tagging. Therefore, it is possible to ask why we developed simple recurrent networks in SCREEN. In this subsection we will provide a detailed comparison of simple recurrent networks and n-gram techniques for the prediction of basic syntactic categories. We chose this task for a detailed comparison since it is currently the most difficult task for a simple recurrent network in SCREEN. So purposefully we did not choose a subtask for which a simple recurrent network had a very high accuracy, but the prediction task since it is more difficult to predict a category compared to disambiguating among categories, for instance. So we chose the difficult prediction with a relatively low network performance in order to be (extremely) fair for the comparison with n-gram techniques.

We are primarily interested in the generalization behavior for new unknown input. Therefore Figure 18 shows the accuracy of the syntactic prediction for the unknown test set. After each word several different syntactic categories can follow and some syntactic categories are excluded. For instance, after a determiner "the" an adjective or a noun can follow: "the short ...", "the appointment", but after a determiner "the" a preposition is implausible to occur and should most probably be excluded. Therefore it is important to know how many categories can be ruled out and Figure 18 shows the relationship between the prediction accuracy and the number of excluded categories for n-grams and our simple recurrent network (as described in Figure 8).

As we can expect, for both techniques, n-grams and recurrent networks, the prediction accuracy is higher if only a few categories have to be excluded and the performance is lower if many categories have to be excluded. However, more interestingly, we can see that simple recurrent networks performed better than 1-grams, 2-grams, 3-grams, 4-grams and 5-grams. Furthermore, it is interesting to note that higher n-grams do not necessarily lead to better performance. For instance, the 4-grams and 5-grams perform worse than 2-grams since they would probably need much larger training sets.

We did the same comparison of n-grams (1-5) and simple recurrent networks also for semantic prediction and received the same result that simple recurrent networks performed better than n-grams. The performance of the best n-gram was often only slightly worse than the performance of the simple recurrent network, which indicates that n-grams are a reasonably useful technique. However, in all comparisons simple recurrent networks per-formed at least slightly better than the best n-grams. Therefore, we used simple recurrent networks as our primary technique for connectionist sequence learning in SCREEN.

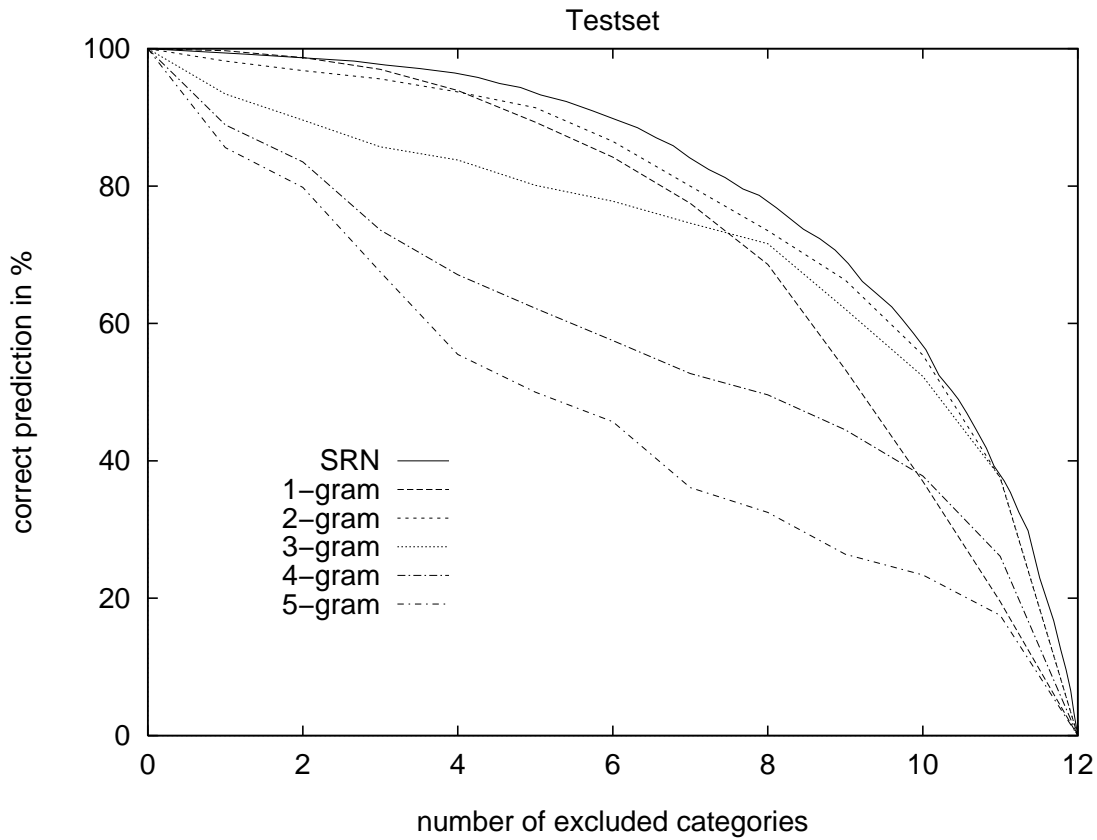

Figure 18: Comparison between simple recurrent network and n-grams

How can we explain this result? N-grams like 2-grams still perform reasonably well for our task and simple recurrent networks are closest to their performance. However, simple recurrent networks perform slightly better since they do not contain a fixed and limited context. In many sequences, the simple recurrent network may primarily use the directly preceding word representation to make a prediction. However, in some exceptions more context is required and the recurrent network has a memory of the internal reduced representation of the preceding context. Therefore, it has the potential to be more flexible with respect to the context size.

N-grams may not perform optimally but they are extremely fast. So the question arises how much time is necessary to compute a new category using new input and the current context for the network. In general our networks differ slightly in size but typically they contain several hundred weights. For a typical representative simple recurrent network with 13 input units, 14 hidden units, 8 output units, and 14 context units, and about 500 weights it takes $10^{-4}s$ on a Sparc Ultra to compute a new category within the whole forward sweep.

Since the techniques for smoothed n-grams basically rely on an efficient table-look-up of precomputed values, of course typical n-gram techniques are still faster. However, due to their fixed-size context they may not perform as well as simple recurrent networks. Furthermore, computing the next possible categories in $10^{-4}s$ is fast enough for our current version of SCREEN. For the sake of an explanation one could argue that SCREEN contains about 10 networks modules and a typical utterance contains 10 words, so a single utterance hypothesis could be performed in $10^{-2}s$. However, different from text tagging, we do not have single sentences but we process word graphs. Depending on the specific utterance, about $10^5$ word hypothesis sequences could be generated and have to be processed. Furthermore there is some book-keeping required for keeping the best word hypotheses, for loading the appropriate networks with the appropriate word hypotheses, etc. The potentially large number of word hypotheses, the additional book-keeping performance, and the number of individual modules for syntax, semantics and dialog processing explain why the total analysis time of the whole unoptimized SCREEN system is in the order of seconds although a single recurrent network performs in the order of $10^{-4}s$.

## 6.2 Improvement in the Hypothesis Space

In this subsection we will analyze to what extent the syntactic and semantic prediction knowledge can be used to improve the best found sentence hypotheses. We illustrate the pruning performance in the hypothesis space by integrating acoustic, syntactic, and semantic knowledge. While the speech recognizer alone provides only acoustic confidence values, SCREEN adds syntactic and semantic knowledge. All these knowledge sources are weighted equally in order to compute a single plausibility value for the current word hypothesis sequence. This plausibility value is used in the speech construction part to prune the hypothesis space and to select the currently best word hypothesis sequences. Several word hypothesis sequences are processed incremental and in parallel. At a given time the $n$ best incremental word hypothesis sequences are kept[11].

The syntactic and semantic plausibility values are based on the basic syntactic and semantic prediction (BAS-SYN-PRE and BAS-SEM-PRE) of the next possible categories for a word and the selection of a preference by the determined basic syntactic respectively semantic category (BAS-SYN-DIS and BAS-SEM-DIS)[12]. The performance of the disambiguation modules is 86%-89% for the test set. For the prediction modules the performance is 72% and 81% for the semantic and syntactic test set, respectively if we want to exclude at least 8 of the 12 possible categories. This performance allows us the computation of a syntactic and semantic plausibility in SYN-SPEECH-ERROR and SEM-SPEECH-ERROR. Based on the combined acoustic, syntactic, and semantic knowledge, first tests on the 184 turns show that the accuracy of the constructed sentence hypotheses of SCREEN could be increased by about 30% using acoustic and syntactic plausibilities and by about 50% using acoustic, syntactic, and semantic plausibilities (Wermter & Weber, 1996a).

---

11. In our experiments low values ($n = 10$) provided the best overall performance.
12. This was explained in more detail in Section 4.3.2

## 6.3 SCREEN's Network Performance and Why the Networks Yield High Accuracy with Little Training

For evaluating the performance of SCREEN's categorization part on the meeting corpus we first show the percentages of correctly classified words for the most important networks for categorization: BAS-SYN-DIS, BAS-SEM-DIS, ABS-SYN-CAT, ABS-SEM-CAT, PHRASE-START. There were 184 turns in this corpus with 314 utterances and 2355 words. 1/3 of the 2355 words and 184 turns was used for training, 2/3 for testing. Usually more data is used for training than testing. In preliminary earlier experiments we had used 2/3 for training and 1/3 for testing. However, the performance on the unknown test set was similar for the 1/3 training set and 2/3 test set. Therefore, we used more testing than training data since we were more interested in the generalization performance for unknown instances in the test set compared to the training performance for known instances.

At first sight, it might seem relatively little data for training. While statistical techniques and information retrieval techniques often work on large texts and individual lexical word items, we need much less material to get a reasonable performance since we work on the syntactic and semantic representations rather than the words. We would like to stress that we use the syntactic and semantic category representations of 2355 words for training and testing rather than the lexical words themselves. Therefore, the category representation requires much less training data than a lexical word representation would have required. As a side effect, also training time was reduced for the 1/3 training set, while keeping the same performance on the 2/3 test set. That is, for training we used category representations from 64 dialog turns, for testing generalization the category representations from the remaining 120 dialog turns.

Table 5 shows the test results for individual networks on the unknown test set. These networks were trained for 3000 epochs with a learning rate of 0.001 and 14 hidden units. This configuration had provided the best performance for most of the network architectures. In general we tested network architectures from 7 to 28 hidden units, learning parameters from 0.1 to 0.0001. As learning rule we used the generalized delta rule (Rumelhart et al., 1986). An assigned output category representation for a word was counted as correct if the category with the maximum activation was the desired category.

| Module | Accuracy on test set |
|---|---|
| BAS-SYN-DIS | 89% |
| BAS-SEM-DIS | 86% |
| ABS-SYN-CAT | 84% |
| ABS-SEM-CAT | 83% |
| PHRASE-START | 90% |
| WORD-ERROR | 94% |
| PHRASE-ERROR | 98% |

Table 5: Performance of the individual networks on the test set of the meeting corpus

The performance for the basic syntactic disambiguation was 89% on the unknown test set. Current syntactic (text-)taggers can reach up to about 95% accuracy on texts. However, there is a big difference between text and speech parsing due to the spontaneous noise in spoken language. The interjections, pauses, repetitions, repairs, new starts and more "ungrammatical" syntactic varieties in our spoken-language domain are reasons why the typical accuracy of other syntactic *text* taggers has not been reached.

On the other hand we see 86% accuracy for the basic semantic disambiguation which is relatively high for semantics. So there is some evidence that the noisy "ungrammatical" variety of spoken language hurts syntax but less semantics. Due to the domain dependence of semantic classifications it is more difficult to compare and explain semantic performance. However, in a different study within the domain of railway interactions we could reach a similar performance (for details see Section 6.6). In all our experiments syntactic results were better than the semantic results, indicating that the syntactic classification was easier to learn and generalize. Furthermore, our syntactic results were close to 90% for noisy spoken language which we consider to be very good in comparison to 95% for more regular text language.

The performance for the abstract categories is somewhat lower than for the basic categories since the evaluation at each word introduces some unavoidable errors. For instance, after "in" the network cannot yet know if a time or location will follow, but has to make an early decision already. In general, the networks perform relatively well on this difficult real-world corpus, given that we did not eliminate any sentence for any reason and took all the spontaneous sentences as they had been spoken.

Furthermore, we use transcripts of spontaneous language for training in the domain of meeting arrangements. Most utterances are questions and answers about dates and locations. This restricts the potential syntactic and semantic constructions, and we certainly benefit from the restricted domain. Furthermore, while some mappings are ambiguous for learning (e.g., a noun can be part of a noun group or a prepositional group) other mappings are relatively unambiguous (e.g., a verb is part of a verb group). We would not expect the same performance on mixed arbitrary domains like the random spoken sentences about various topics from passers-by in the city. However, the performance in somewhat more restricted domains can be learned in a promising manner (for a transfer to a different domain see Section 6.6). So there is some evidence that simple recurrent networks can provide good performance using small training data from a restricted domain.

## 6.4 SCREEN's Overall Output Performance

While we just described the individual network performance, we will now focus on the performance of the running system. The performance in the running SCREEN system has to be different from the performance of the individual networks for a number of reasons. First, the individual networks are trained separately in order to support a modular architecture. In the running SCREEN system, however, connectionist networks receive their input from other underlying networks. Therefore, the actual input to a connectionist network in the running SCREEN system may also differ from the original training and test sets. Second, the spoken sentences may contain errors like interjections or word repairs. These have to be part of the individual network training, but the running SCREEN system is able to detect and correct certain interjections, word corrections and phrase corrections. Therefore, system and network performance differ at such disfluencies. Third, if we want to evaluate the performance of abstract semantic categorization and abstract syntactic categorization we are particularly interested in certain sentence parts. For abstract syntactic categorization, e.g., the detection of a prepositional phrase, we have to consider that the beginning of a phrase with its significant function word, e.g., preposition, should be the most important location for syntactic categorization. In contrast, for abstract semantic categorization, the content word at the end of a phrase group, directly before the next phrase start, is most important.

| | |
|---|---|
| Correct flat syntactic output representation | 74% |
| Correct flat semantic output representation | 72% |

Table 6: Overall syntactic and semantic accuracy of the running SCREEN system on the unknown test set of the meeting corpus

As we should expect based on the explanation in the previous paragraph, the overall accuracy of the output of the complete running system should be lower than the performance of the individual modules. In fact, this is true and Table 6 shows the overall syntactic and semantic phrase accuracy of the running SCREEN system. 74% of all assigned syntactic phrase representations of the unknown test set are correct and 72% of all assigned semantic phrase representations. The slight performance drop can be partially explained by the more uncertain input from other underlying networks which themselves are influenced by other networks. On the other hand, in some cases the various decisions by different modules (e.g. the three modules for lexical, syntactic and semantic category equality of two words) can be combined in order to clean up some errors (e.g. a wrong decision by one single module). In general, given that the 120 dialog turns of the test set were completely unrestricted, unknown real-world and spontaneous language turns, we believe that the overall performance is quite promising.

## 6.5 SCREEN's Overall Performance for an Incomplete Lexicon

One important property of SCREEN is its robustness. Therefore, it is an interesting question how SCREEN would behave if it could only receive incomplete input from its lexicon. Such situations are realistic since speakers could use new words which a speech recognizer has not seen before. Furthermore, we can test the robustness of our techniques. While standard context-free parsers usually cannot provide an analysis if words are missing from the lexicon, SCREEN would not break on missing input representations, although of course we have to expect that the overall classification performance must drop if less reliable input is provided.

In order to test such a situation under the controlled influence of removing items from the lexicon, we first tested a scenario where we randomly eliminated 5% of the syntactic and semantic lexicon representations. If a word was unknown, SCREEN used a single syntactic and single semantic average default vector instead. This average default vector contained the normalized frequency of each syntactic respectively semantic category across the lexicon.

Even without 5% of all lexicon entries all utterances could still be analyzed. So SCREEN does not break for missing word representations but attempts to provide an analysis as good

| Correct flat syntactic output representation | 72% |
| Correct flat semantic output representation | 67% |

Table 7: Overall syntactic and semantic accuracy of the running SCREEN system for the meeting corpus on the unknown test set after 5% of all lexicon entries were eliminated

as possible. As expected, Table 7 shows a performance drop for the overall syntactic and semantic accuracy. However, compared to the 74% and 72% performance for the complete lexicon (see Table 6) we still find that 72% of the syntactic output representations and 67% of the semantic output representations are correct after eliminating 5% of all lexicon entries.

| Correct flat syntactic output representation | 70% |
| Correct flat semantic output representation | 67% |

Table 8: Overall syntactic and semantic accuracy of the running SCREEN system for the meeting corpus on the unknown test set after 10% of all lexicon entries were eliminated

In another experiment we eliminated 10% of all syntactic and semantic lexicon entries. In this case, the syntactic accuracy was still 70% and the semantic accuracy was 67%. Eliminating 10% of the lexicon led to a syntactic accuracy reduction of only 4% (74% versus 70%) and a semantic accuracy reduction of 5% (72% versus 67%). In general we see that in all our experiments the percentage of accuracy reduction was much less than the percentage of eliminated lexicon entries demonstrating SCREEN's robustness for working with an incomplete lexicon.

## 6.6 Comparison with the Results in a New Different Domain

In order to compare the performance of our techniques, we will also show results from experiments with a different spoken Regensburg Train Corpus. Our intention cannot be to describe the experiments in this domain at the same level of detail as we have done for our Blaubeuren Meeting Corpus in this paper. However, we will provide a summary in order to provide a point of reference and comparison for our experiments on the meeting corpus. This comparison serves as another additional possibility to judge our results for the meeting corpus.

As a different domain we chose 176 dialog turns at a railway counter. People ask questions and receive answers about train connections. A typical utterance is: "Yes I need eh a a sleeping car PAUSE from PAUSE Regensburg to Hamburg". We used exactly the same SCREEN communication architecture to process spoken utterances from this domain: the same architecture was used, 1/3 of the dialog turns was used for training, 2/3 for testing on unseen unknown utterances. For syntactic processing, we even used exactly the same network structure, since we did not expect much syntactic differences between the two domains. Only for semantic processing we retrained the semantic networks. Different categories had to be used for semantic classification, in particular for actions. While actions about meetings (e.g., visit, meet) were predominant in the meeting corpus, actions about selecting connections (e.g., choose, select) were important in the train corpus (Wermter & Weber, 1996b). Just to give the reader an impression of the portability of SCREEN, we would estimate that 90% of the original human effort (system architecture, networks) could be used in this new domain. Most of the remaining 10% were needed for the necessary new semantic tagging and training in the new domain.

| Module | Accuracy on test set |
|---|---|
| BAS-SYN-DIS | 93% |
| BAS-SEM-DIS | 84% |
| ABS-SYN-CAT | 85% |
| ABS-SEM-CAT | 77% |
| PHRASE-START | 89% |
| WORD-ERROR | 94% |
| PHRASE-ERROR | 98% |

Table 9: Performance of the individual networks on the test set in the train corpus

Table 9 shows the performance on the test set in the train corpus. If we compare our results in the meeting corpus (Table 5) with these results in the train corpus we see in particular that the abstract syntactic processing is almost the same in the meeting corpus (84% in Table 5 compared to 85% in Table 9) but the abstract semantic processing is better in the meeting corpus (83% in Table 5 compared to 77% in Table 9). Other modules dealing with explicit robustness for repairs (phrase start, word repair errors, phrase repair errors) show almost the same performance (90% vs 89%, 94% vs 94%, 98% vs 98%).

| | |
|---|---|
| Correct flat syntactic output representation | 76% |
| Correct flat semantic output representation | 64% |

Table 10: Overall syntactic and semantic accuracy of the running SCREEN system on the unknown test set of a different train corpus

As a comparison we summarize here the overall performance for this different train domain. Table 10 shows that SCREEN has about the same syntactic performance in the two domains (compare with Table 6). So in this different domain we can essentially confirm our previous results for syntactic processing performance (74% vs. 76%). However, semantic processing appears to be harder in the train domain since the performance of 64% is lower than the 72% in the meeting domain. However, semantic processing, semantic tagging or semantic classification is often found to be much harder than syntactic processing in general, so that the difference is still within the range of usual performance differences in syntax and semantics. Since semantic categories like agents, locations, and time expressions are about the same in these two domains the more difficult action categorization is mainly responsible for this difference in semantic performance between the two domains.

In general the transfer from one domain to another only requires a limited amount of hand-modeling. Of course, syntactic and semantic categories have to be specified for the lexicon and the transcripts. These syntactically or semantically tagged transcript sentences are the direct basis for generating the training sets for the networks. Generating these trainings sets is the main manual effort while transferring the system to a new domain. After the generation of the training sets has been performed the training of the networks can proceed automatically. The training of a typical single recurrent network takes in the order of a few hours. So much less manual work is required than for transferring a standard symbolic parser to a new domain and generating a new syntactic and semantic grammar.

## 6.7 An Illustrative Comparison Argument Based on a Symbolic Parser

We have made the point that SCREEN's learned flat representations are more robust than hand-coded deeply structured representations. Here we would like to elaborate this point with a compelling illustrative argument. Consider different variations of sentence hypotheses from a speech recognizer in Figure 19: 1. A correct sentence hypothesis: "Am sechsten April bin ich außer Hause" ("On 6th April am I out of home") and 2. A partially incorrect sentence hypothesis: "Am sechsten April ich ich außer Hause" ("On 6th April I I out of home"). Focusing on the syntactic analysis, we used an existing chart parser and an existing grammar which had been used extensively for other real-world parsing up to the sentence level (Wermter, 1995). The only necessary significant adaptation was the addition of a rule $NG \rightarrow U$ for pronouns, which had not been part of the original grammar. This rule states that a pronoun U (e.g., "I") can be a noun group (NG).

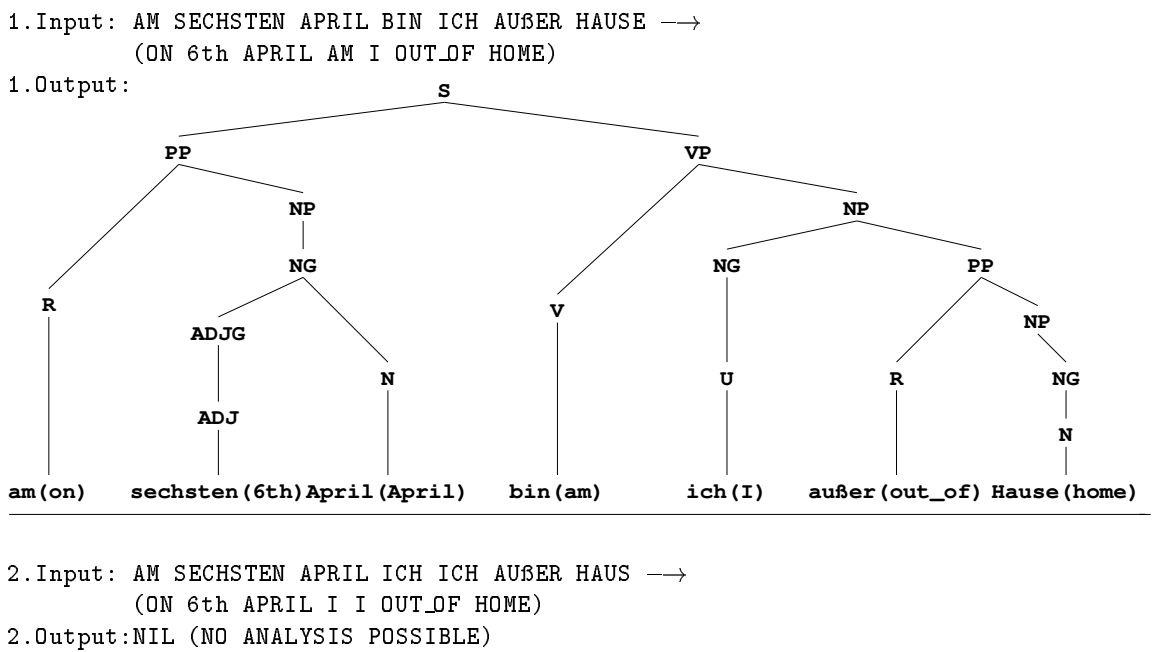

Figure 19: Two sentence hypotheses from a speech recognizer. The first hypothesis can be analyzed, the second partially incorrect hypothesis cannot be analyzed anymore by the symbolic parser.

If we run the first sentence hypothesis through the symbolic context-free parser we receive the desired syntactic analysis shown in Figure 19, but if we run the second slightly incorrect sentence hypothesis through the parser we do not receive any analysis (The syntactic category abbreviations in Figure 19 are used in the same manner as throughout the paper (see Table 1-4); furthermore and as usual, "S" stands for sentence, "ADJG" for adjective group, "NP" for complex nominal phrase, "VP" for verb phrase. The literal English translations are shown in brackets).

The reason why the second sentence hypothesis could not be parsed by the context-free chart parser was that the speech recognizer generated incorrect output. There is no verb in the second sentence hypothesis and there is an additional pronoun "I". Such mistakes occur rather frequently based on the imperfectness of current speech recognition technology. Of course one could argue that the grammar should be relaxed and made more flexible to deal with such mistakes. However, the more rules for fault detection are integrated into the grammar or the parser the more complicated the grammar or the parser. Even more important, it is impossible to predict all possible mistakes and integrate them into a symbolic context-free grammar. Finally, relaxing the grammar for dealing with mistakes by using explicit specific rules also might lead to other additional mistakes because the grammar now has to be extremely underspecified.

As we have shown, for instance in Figure 17, SCREEN does not have problems dealing with such speech recognizer variations and mistakes. The main difference between a standard context-free symbolic chart parser analysis and SCREEN's analysis is that SCREEN has learned to provide a flat analysis under noisy conditions but the context-free parser has been hand-coded to provide a more structural analysis. It should be emphasized here that we do not make an argument against structural representations per se and in general. The more structure that can be provided the better, particularly for tasks which require structured world knowledge. However, if robustness is a major concern, as it is for lower syntactic and semantic spoken-language analysis, a learned flat analysis provides more robustness.

## 6.8 Comparisons with Related Hybrid Systems

Recently, connectionist networks have received a lot of attention as computational learning mechanisms for written language processing (Reilly & Sharkey, 1992; Miikkulainen, 1993; Feldman, 1993; Barnden & Holyoak, 1994; Wermter, 1995). In this paper however, we have focused on the examination of hybrid connectionist techniques for *spoken* language processing. In most previous approaches to speech/language processing processing was often sequential. That is, one module like the speech recognizer or the syntactic analyzer completed its work before the next module like a semantic analyzer started to work. In contrast, SCREEN works incrementally which allows the system (1) to have modules running in par-allel, (2) to integrate knowledge sources very early, and (3) to compute the analysis more similar to humans since humans start to process sentences before they may be completed.

We will now compare our approach to related work and systems. A head-to-head comparison with a different system is difficult based on different computer environments and whether systems can be accessed and adapted easily for the same input. Furthermore, different systems are typically used for different purposes with different language corpora, grammars, rules, etc. However, we have made an extensive effort for a fair conceptual comparison.

PARSEC (Jain, 1991) is a hybrid connectionist system which is embedded in a larger speech translation effort JANUS (Waibel et al., 1992). The input for PARSEC is sentences, the output is case role representations. The system consists of several connectionist modules with associated symbolic transformation rules for providing transformations suggested by the connectionist networks. While it is PARSEC's philosophy to use connectionist networks for triggering symbolic transformations, SCREEN uses connectionist networks for the transformations themselves. It is SCREEN's philosophy to use connectionist networks wherever possible and symbolic rules only where they are necessary.

We found symbolic processing particularly useful for simple known tests (like lexical equality) or for complex control tasks of the whole system (when does a module communicate to which other module). Much of the actual transformational work can be done by trained connectionist networks. This is in contrast to the design philosophy in PARSEC where connectionist modules provide control knowledge which transformation should be performed. Then the selected transformation is actually performed by a symbolic procedure. So SCREEN uses connectionist modules for transformations and a symbolic control, while PARSEC uses connectionist modules for control and symbolic procedures for the transformations.

Different from SCREEN, PARSEC receives sentence hypotheses either as sentence transcripts or as N-best hypotheses from the JANUS system. Our approach receives incremental word hypotheses which are used in the speech construction part to build sentence hypotheses. This part is also used to prune the hypothesis space and to determine the best sentence hypotheses. So during the flat analysis in SCREEN the semantic and syntactic plausibilities of a partial sentence hypothesis can still influence which partial sentence hypotheses are processed.

For PARSEC and for SCREEN a modular architecture was tested which has the advantage that each connectionist module has to learn a relatively easy subtask. In contrast to the development of PARSEC it is our experience that modularity requires less training time. Furthermore, some modules in SCREEN are able to work independently from each other and in parallel. In addition to syntactic and semantic knowledge, PARSEC can make use of prosodic knowledge while SCREEN currently does not use prosodic hints. On the other hand, SCREEN also contains modules for learning dialog act assignment while such modules are currently not part of PARSEC. Learning dialog act processing is important for determining the intended meaning of an utterance (Wermter & Löchel, 1996).

Recent further extensions based on PARSEC provide more structure and use annotated linguistic features (Buø et al., 1994). The authors state that they "implemented (based on PARSEC) a connectionist system" which should approximate a shift reduce parser. This connectionist shift-reduce parser substantially differs from the original PARSEC architecture.

We will refer to it as the "PARSEC extension". This PARSEC extension labels a complete sentence with its first level categories. These first level categories are input again to the same network in order to provide second level categories for the complete sentence and so on, until at the highest level the sentence symbol can be added.

Using this recursion step the PARSEC extension can provide deeper and more structural interpretations than SCREEN currently does. However, this recursion step and the construction of the structure also have their price. First, labels like NP for a noun phrase have to be defined as lexical items in the lexicon. Second, and more important, the complete utterance is labeled with the $n$-th level categories before processing with the $n+1$-th level categories starts. Therefore several parses (e.g., 7 for the utterance "his big brother loved himself") through the utterance are necessary. This means that this recent PARSEC extension is more powerful than SCREEN and the original PARSEC system by Jain with respect to the opportunity to provide deeper and more structural interpretations. However, at the same time this PARSEC extension looses the possibility to process utterances in an incremental manner. However, incrementality is a very important property in spoken-language processing and in SCREEN. Besides the fact that humans process language in an incremental left-to-right manner, this also allows SCREEN to prune the search space of incoming word hypotheses very early.

Comparing PARSEC and SCREEN, PARSEC aims more at supporting symbolic rules by using symbolic transformations (triggered by connectionist networks) and by integrating linguistic features. Currently, the linguistic features in the recent PARSEC extension (Buø et al., 1994) provide more structural and morphological knowledge than SCREEN does. Therefore, currently it appears to be easier to integrate the PARSEC extension into larger systems of high level linguistic processing. In fact, PARSEC has been used in the context of the JANUS framework. On the other hand, SCREEN aims more at robust and incremental processing by using a word hypothesis space, specific repair modules, and more flat representations. In particular, SCREEN emphasizes more the robustness of spoken-language processing, since it contains explicit repair mechanisms and implicit robustness. Explicit robustness covers often occurring errors (interjections, pauses, word and phrase repairs) in explicit modules, while other less predictable types of errors are only supported by the implicit similarity-based robustness from the connectionist networks themselves. In general, the representations generated by the extension of PARSEC provide better support for deeper structures than SCREEN, but SCREEN provides better support for incremental robust processing. In a more recent extension based on PARSEC called FeasPar, the overall parsing performance was a syntactic and semantic feature accuracy of 33.8%. Although additional improvements can be shown using subsequent search techniques on the parsing results, we did not consider such subsequent search techniques for better parses since they would violate incremental processing (Buø, 1996). Without using subsequent search techniques SCREEN reaches an overall semantic and syntactic accuracy between 72% and 74% as shown in Table 6. However it should be pointed out, that SCREEN and FeasPar use different input sentences, features and architectures.

Besides PARSEC also the BeRP and TRAINS systems focus on hybrid spoken-language processing. BeRP (Berkeley Restaurant Project) is a current project which employs multiple different representations for speech/language analysis (Wooters, 1993; Jurafsky et al., 1994, 1994b). The task of BeRP is to act as a knowledge consultant for giving advice about choos-ing restaurants. There are different components in BeRP: The feature extractor receives digitized acoustic data and extracts features. These features are used in the connectionist phonetic probability estimation. The output of this connectionist feedforward network is used in a Viterbi decoder which uses a multiple pronunciation lexicon and different language models (e.g. bigram, hand-coded grammar rules). The output of the decoder are word strings which are transformed into database queries by a stochastic chart parser. Finally, a dialog manager controls the dialog with the user and can ask questions.

BeRP and SCREEN have in common the ability to deal with errors from humans and from the speech recognizer as well as a relatively flat analysis. However, for reaching this robustness in BeRP a probabilistic chart parser is used to compute all possible fragments at first. Then, an additional fragment combination algorithm is used for combining these fragments so that they cover the greatest number of input words. Different from this sequential process of first computing all fragments of an utterance and then combining the fragments, SCREEN uses incremental processing and desirably provides the best possible interpretation. In this sense SCREEN's language analysis is weaker but more general. SCREEN's analysis will never break and produce the best possible interpretation for all noisy utterances. This strategy may be particularly useful for incremental translation. On the other hand, BeRP's language analysis is stronger but more restricted. BeRP's analysis may stop at the fragment level if there are contradictory fragments. This strategy may be particularly useful for question answering where additional world knowledge is necessary and available.

TRAINS is a related spoken-language project for building a planning assistant who can reason about time, actions, and events (Allen, 1995; Allen et al., 1995). Because of this goal of building a general framework for natural language processing and planning for train scheduling, TRAINS needs a lot of commonsense knowledge. In the scenario, a person interacts with the system in order to find solutions for train scheduling in a cooperative manner. The person is assumed to know more about the goals of the scheduling while the system is supposed to have the details of the domain. The utterance of a person is parsed by a syntactic and semantic parser. Further linguistic reasoning is completed by modules for scoping and reference resolution. After the linguistic reasoning, conversation acts are determined by a system dialog manager and responses are generated based on a template-driven natural language generator. Performance phenomena in spoken language like repairs and false starts can currently be dealt with already (Heeman & Allen, 1994b, 1994a). Compared to SCREEN, the TRAINS project focuses more on processing spoken language at an in-depth planning level. While SCREEN uses primarily a flat connectionist language analysis, TRAINS uses a chart parser with a generalized phrase structure grammar.

## 7. Discussion

First we will focus on what has been learned for *processing spoken-language processing*. When we started the SCREEN project, it was not predetermined whether a deep analysis or a flat screening analysis would be particularly appropriate for robust analysis of spoken sentences. A deep analysis with highly structured representations is less appropriate since the unpredictable faulty variations in spoken language limit the usefulness of deep structured knowledge representations much more than it is the case for written language. Deep interpretations and very structured representations - as for instance possible with HPSG grammars for text processing - make a great deal of assumptions and predictions which do not hold for faulty spoken language. Furthermore, we have learned that for generating a semantic and syntactic representation we do not even need to use a deep interpretation for certain tasks. For instance, for translating between two languages it is not necessary to disambiguate all prepositional phrase attachment ambiguities since during the process of translation these disambiguations may get ambiguous again in the target language.

However, we use some structure at the level of words and phrases for syntax and semantics respectively. We learned that a single flat semantics level rather than the four flat syntax and semantics levels is not sufficient since syntax is necessary for detecting phrase boundaries. One could argue that one syntactic abstract phrase representation and one abstract semantic phrase representation may be enough. However, we found that the basic syntactic and semantic representations at the word level make the task easier for the subsequent abstract analysis at the phrase level. Furthermore, the basic syntactic and semantic representations are necessary for other tasks as well, for instance for the judgment of the plausibility of a sequence of syntactic and semantic categories. This plausibility is used as a filter for finding good word hypothesis sequences. Therefore, we argue that for processing faulty spoken language - for a task like sentence translation or question answering - we need much less structured representations as are typically used in well-known parsers but we need more structured representations than those of a single-level tagger.

In some of our previous work we had made early experiences with related connectionist networks for *analyzing text phrases*. Moving from analyzing text phrases to analyzing unrestricted spoken utterances, there are tremendous differences in the two tasks. We found that the phrase-oriented flat analysis used in SCAN (Wermter, 1995) is advantageous in principle for *spoken-language analysis* and the phrase-oriented analysis is common to learning text and speech processing. However, we learned that spoken-language analysis needs a much more sophisticated architecture. In particular, since spoken language contains many unpredictable errors and variations, fault tolerance and robustness are much more important. Connectionist networks have an inherent implicit robustness based on their similarity-based processing in gradual numerical representations. In addition, we found that for some classes of relatively often occurring mistakes, there should be some explicit robustness provided by machinery for interjections, word and phrase repairs. Furthermore, the architecture has to consider the processing of a potentially large number of competing word hypothesis sequences rather than a single sentence or phrase for text processing.

Now, we will focus on what has been learned about *connectionist and hybrid architectures*. In the beginning we did not predetermine whether connectionist methods would be particularly useful for control or for individual modules or for both. However, during the development of the SCREEN system it became clear that for the general task of *spoken* language understanding, individual subtasks like syntactic analysis had to be very fault-tolerant because of the "noise" in spoken language, due both to humans and to speech-recognizers as well. Especially unforeseeable variations often occur in spontaneously spoken language and cannot be predefined well in advance as symbolic rules in a general manner. This fault-tolerance at the task level could be supported particularly well by the inherent fault-tolerance of connectionist networks for individual tasks and the support of inductive learning algorithms. So we learned that for a flat robust understanding of spoken-language connectionist networks are particularly effective within individual subtasks.

There has been quite a lot of work on control in connectionist networks. However, in many cases these approaches have concentrated on control in single networks. Only recently there has been more work on control in modular architectures (Sumida, 1991; Jacobs et al., 1991b; Jain, 1991; Jordan & Jacobs, 1992; Miikkulainen, 1996). For instance, in the approach by Jacobs and Jordan (Jacobs et al., 1991b; Jordan & Jacobs, 1992), task knowledge and control knowledge are learned both. Task knowledge is learned in individual task networks, and higher control networks are responsible for learning when a single task network is responsible for producing the output. Originally it was an open question whether a connectionist control would be possible for processing spoken language. While automatic modular task decomposition (Jacobs et al., 1991a) can be done for simple forms of function approximation, more complex problems like understanding spoken language in real-world environments still need designer-based modular task decomposition for the necessary tasks.

We learned that connectionist control in an architecture with *a lot of* modules and subtasks currently seems to be beyond the capabilities of current connectionist networks. It has been shown that connectionist control is possible for a limited number of connectionist modules (Miikkulainen, 1996; Jain, 1991). For instance Miikkulainen shows that a connectionist segmenter and a connectionist stack can control a parser to analyze embedded clauses. However, the communication paths still have to be very restricted within these three modules. Especially for a real-world system for spoken-language understanding from speech, over syntax, semantics to dialog processing for translation it is extremely difficult to learn to coordinate the different activities, especially for a large parallel stream of word hypothesis sequences. We believe that it may be possible in the future, however currently connectionist control in SCREEN is restricted to the detection of certain hesitations phenomena like corrections.

Considering flat screening analysis of spoken language and hybrid connectionist techniques together, we have developed and followed a general guideline (or *design philosophy*) of using as little knowledge as necessary while getting as far as possible using connectionist networks wherever possible and symbolic representations where necessary. This guideline led us to (1) a flat but robust representation of spoken-language analysis and to (2) the use of hybrid connectionist techniques which support the task by the choice of the possibly most appropriate knowledge structure. Many hybrid systems contain just a small portion of connectionist representations in addition to many other modules, e.g. BeRP (Wooters, 1993; Jurafsky et al., 1994, 1994b), JANUS (Waibel et al., 1992), TRAINS (Allen, 1995; Allen et al., 1995). In contrast, most of the important subtasks in SCREEN are performed directly by many connectionist networks.

Furthermore, we have learned that flat syntactic and semantic representations could give surprisingly good training and test results when trained and tested with a medium corpus of about 2300 words in the 184 dialog turns. These good results are mostly due to the learned internal weight representation and the local context which adds sequentiality to the category assignments. Without the internal weight representation of the preceding context the syntactic and semantic categorization does not perform equally well, so the choice of recurrent networks is crucial for many sequential category assignments. Therefore these networks and techniques hold potential especially for such medium-size domains where a restricted amount of training material is available. While statistical techniques are often used for very large data sets, but do not work well for medium data sets, the connectionist techniques we used work well for medium-size domains.

The used techniques can be ported to different domains and be used for different purposes. Even if different sets of categories would have to be used the learning networks are able to extract these syntactic regularities automatically. Besides the domain of arranging business meetings we have also ported SCREEN to the domain of interactions at a railway counter with comparable syntactic and semantic results. These two domains differed primarily in their semantic categories, while the syntactic categories (and networks) of SCREEN could be used directly.

SCREEN has the potential for scaling up. In fact, based on the imperfect output of a speech recognizer, several thousand sentence hypotheses have already been processed. If new words are to be processed, their syntactic and semantic basic categories are simply entered into the lexicon. The structure of individual networks does not change, new units do not have to be added and therefore the networks do not have to be retrained.

The amount of hand-coding is restricted primarily to the symbolic control of the module interaction and to the labeling of the training material for the individual networks. When we changed the domain to railway counter interactions, we could use the identical control, as well as the syntactic networks. Only the semantic networks had to be retrained due to the different domain.

So far we have focused on supervised learning in simple recurrent networks and feedforward networks. Supervised learning still requires a training set and some manual labeling work still has to be done. Although especially for medium size corpora labeling examples is easier than for instance designing complete rule bases it would be nice to automate the knowledge acquisition even further. Currently we plan to build a more sophisticated lexicon component which will provide support for automatic lexicon design (Riloff, 1993) and dynamic lexicon entry determination using local context (Miikkulainen, 1993).

Furthermore, SCREEN could be expanded at the speech construction and evaluation part. The syntactic and semantic hypotheses could be used for more interaction with the speech recognizer. Currently syntactic and semantic hypotheses from the speech evaluation part are used to exclude unlikely word hypothesis sequences from the language modules. However, these hypotheses by the connectionist networks for syntax and semantics - in particular the modules of basic syntactic and semantic category prediction - could also be used directly into the process of recognition in the future in order to provide more syntactic and semantic feedback to the speech recognizer at an early stage. Besides syntax and semantics, cue phrases, stress and intonation could provide additional knowledge for speech/language processing (Hirschberg, 1993; Gupta & Touretzky, 1994). These issues will be additional major efforts for the future.

## 8. Conclusions

We have described the underlying principles, the implemented architecture, and the evaluation of a new screening approach for *learning the analysis of spoken language*. This work makes a number of original contributions to the fields of artificial intelligence and advances the state of the art in several perspectives: From the perspective of symbolic and connectionist design we argue for a hybrid solution, where connectionist networks are used wherever they are useful but symbolic processing is used for control and higher level analysis. Furthermore, we have shown that recurrent networks provided better syntactic and semantic prediction results than 1-5 grams. From the perspective of connectionist networks alone, we have demonstrated that connectionist networks can in fact be used in *real-world* spoken-language analysis. From the perspective of natural language processing we argue that hybrid system design is advantageous for integrating speech and language since lower speech-related processing is supported by fault-tolerant learning in connectionist networks and higher processing and control is supported by symbolic knowledge structures. In general, these properties support parallel rather than sequential, learned rather than coded, fault-tolerant rather than strict processing of spoken language.

The main result of this paper is that learned flat representations support robust processing of spoken language better than in-depth structured representations and that connectionist networks provide a fault-tolerance to reach this robustness. Due to the noise in spontaneous language (interjections, pauses, repairs, repetitions, false starts, ungrammaticalities, and also additional false word hypotheses by a speech recognizer) complex structured possibly recursive representations often cannot be computed using standard symbolic representations like context-free parsers. On the other hand, there are tasks like information extraction from of spoken language which may not need an in-depth structured representation. We believe our hybrid connectionist techniques have considerable potential for such tasks, for instance for information extraction in restricted but noisy spoken-language domains. While an in-depth understanding like inferencing for story interpretation needs complex structured representations, a shallow understanding for instance for information extraction in noisy speech language environments will benefit from flat, robust and learned representations.

## Acknowledgements

This research was funded by the German Federal Ministry for Research and Technology (BMBF) under Grant #01IV101A0 and by the German Research Association (DFG) under Grant DFG Ha 1026/6-3, and Grant DFG We 1468/4-1. We would like to thank S. Haack, M. Löchel, M. Meurer, U. Sauerland, and M. Schrattenholzer for their work on SCREEN; as well as David Bean, Alexandra Klein, Steven Minton, Johanna Moore, Ellen Riloff and five anonymous referees for comments on earlier versions of this paper.

## References

Allen, J. F. (1995). The TRAINS-95 parsing system: A user's manual. Tech. rep. TRAINS TN 95-1, University of Rochester, Computer Science Department.

Allen, J. F., Schubert, L. K., Ferguson, G., Heeman, P., Hwang, C. H., Kato, T., Light, M., Martin, N. G., Miller, B. W., Poesio, M., & Traum, D. R. (1995). The TRAINS project: A case study in building a conversational planning agent. *Journal of Experimental and Theoretical AI, 7*, 7–48.

Barnden, J. A., & Holyoak, K. J. (Eds.). (1994). *Advances in Connectionist and Neural Computation Theory*, Vol. 3., Ablex Publishing Corporation.

Buø, F. D. (1996). *FeasPar - A Feature Structure Parser Learning to Parse Spontaneous Speech*. Ph.D. thesis, University of Karlsruhe, Karlsruhe, FRG.

Buø, F. D., Polzin, T. S., & Waibel, A. (1994). Learning complex output representations in connectionist parsing of spoken language. In *Proceedings of the International Conference on Acoustics, Speech and Signal Processing*, Vol. 1, pp. 365–368, Adelaide, Australia.

Charniak, E. (1993). *Statistical Language Learning*. MIT Press, Cambridge, MA.

Dyer, M. G. (1983). *In-Depth Understanding: A Computer Model of Integrated Processing for Narrative Comprehension*. MIT Press, Cambridge, MA.

Elman, J. L. (1990). Finding structure in time. *Cognitive Science, 14*(2), 179–211.

Faisal, K. A., & Kwasny, S. C. (1990). Design of a hybrid deterministic parser. In *Proceedings of the 13$^{th}$ International Conference on Computational Linguistics*, pp. 11–16, Helsinki, Finnland.

Feldman, J. A. (1993). Structured connectionist models and language learning. *Artificial Intelligence Review, 7*(5), 301–312.

Geutner, P., Suhm, B., Buø, F.-D., Kemp, T., Mayfield, L., McNair, A. E., Rogina, I., Schultz, T., Sloboda, T., Ward, W., Woszczyna, M., & Waibel, A. (1996). Integrating different learning approaches into a multilingual spoken language translation system. In Wermter, S., Riloff, E., & Scheler, G. (Eds.), *Connectionist, Statistical and Symbolic Approaches to Learning for Natural Language Processing*, pp. 117–131, Springer, Heidelberg.

Gupta, P., & Touretzky, D. S. (1994). Connectionist models and linguistic theory: Investigations of stress systems in language. *Cognitive Science, 18*(1), 1–50.

Hauenstein, A., & Weber, H. H. (1994). An investigation of tightly coupled time synchronous speech language interfaces using a unification grammar. In *Proceedings of the 12$^{th}$ National Conference on Artificial Intelligence Workshop on the Integration of Natural Language and Speech Processing*, pp. 42–49, Seattle, Washington.

Heeman, P. A., & Allen, J. (1994a). Detecting and correcting speech repairs. In *Proceedings of the 32$^{nd}$ Annual Meeting of the Association for Computational Linguistics*, pp. 295–302, Las Cruces, NM.

Heeman, P. A., & Allen, J. (1994b). Tagging speech repairs. In *Proceedings of the Human Language Technology Workshop*, pp. 187–192, Plainsboro, NJ.

Hendler, J. A. (1989). Marker-passing over microfeatures: Towards a hybrid symbolic/connectionist model. *Cognitive Science, 13*(1), 79–106.

Hirschberg, J. (1993). Pitch accent in context: Predicting intonational prominence from text. *Artificial Intelligence, 63,* 305–340.

Jacobs, R. A., Jordan, M. I., & Barto, A. G. (1991a). Task decomposition through competition in a modular connectionist architecture: The what and where vision tasks. *Cognitive Science, 15,* 219–250.

Jacobs, R. A., Jordan, M. I., Nowlan, S. J., & Hinton, G. E. (1991b). Adaptive mixtures of local experts. *Neural Computation, 3*(1), 79–87.

Jain, A. N. (1991). Parsing complex sentences with structured connectionist networks. *Neural Computation, 3*(1), 110–120.

Jordan, M. I., & Jacobs, R. A. (1992). Hierarchies of adaptive experts. In Moody, J. E., Hanson, S. J., & Lippmann, R. R. (Eds.), *Advances in Neural Information Processing Systems 4,* pp. 985–992, Morgan Kaufmann, San Mateo, CA.

Jurafsky, D., Wooters, C., Tajchman, G., Segal, J., Stolcke, A., Fosler, E., & Morgan, N. (1994a). The Berkeley Restaurant Project. In *Proceedings of the International Conference on Speech and Language Processing.* pp. 2139-2142, Yokohama, Japan.

Jurafsky, D., Wooters, C., Tajchman, G., Segal, J., Stolcke, A., & Morgan, N. (1994b). Integrating experimental models of syntax, phonology, and accent/dialect in a speech recognizer. An investigation of tightly coupled time synchronous speech. In *Proceedings of the 12th National Conference on Artificial Intelligence Workshop on the Integration of Natural Language and Speech Processing,* pp. 107–115, Seattle, Washington.

Medsker, L. R. (1994). *Hybrid Neural Network and Expert Systems.* Kluwer Academic Publishers, Boston.

Mellish, C. S. (1989). Some chart-based techniques for parsing ill-formed input. In *Proceedings of the 27th Annual Meeting of the Association for Computational Linguistics,* pp. 102–109, Vancouver, Canada.

Menzel, W. (1994). Parsing of spoken language under time constraints. In Cohn, A. G. (Ed.), *Proceedings of the 11th European Conference on Artificial Intelligence,* pp. 561–564, Amsterdam.

Miikkulainen, R. (1993). *Subsymbolic Natural Language Processing. An integrated model of scripts, lexicon and memory.* MIT Press, Cambridge, MA.

Miikkulainen, R. (1996). Subsymbolic case-role analysis of sentences with embedded clauses. *Cognitive Science, 20,* 47–73.

Nakatani, C., & Hirschberg, J. (1993). A speech-first model for repair detection and correction. In *Proceedings of the 31st Annual Meeting of the Association for Computational Linguistics,* pp. 46–53 Columbus, Ohio.

Reilly, R. G., & Sharkey, N. E. (Eds.). (1992). *Connectionist Approaches to Natural Language Processing.* Lawrence Erlbaum Associates, Hillsdale, NJ.

Riloff, E. (1993). Automatically constructing a dictionary for information extraction tasks. In *Proceedings of the 11th National Conference on Artificial Intelligence*, pp. 811–816, Washington, DC.

Rumelhart, D. E., Hinton, G. E., & Williams, R. J. (1986). Learning internal representations by error propagation. In Rumelhart, D. E., McClelland, J. L., & The PDP research group (Eds.), *Parallel Distributed Processing*, Vol. 1., pp. 318–362. MIT Press, Cambridge, MA.

Sauerland, U. (1996). Konzeption und Implementierung einer Speech/Language Schnittstelle. Master's thesis, University of Hamburg, Computer Science Department, Hamburg, FRG.

Sumida, R. A. (1991). Dynamic inferencing in parallel distributed semantic networks. In *Proceedings of the 13th Annual Meeting of the Cognitive Science Society*, pp. 913–917, Boston, Chicago.

Sun, R. (1994). *Integrating Rules and Connectionism for Robust Common Sense Reasoning*. Wiley and Sons, New York.

von Hahn, W., & Pyka, C. (1992). System architectures for speech understanding and language processing. In Heyer, G., & Haugeneder, H. (Eds.), *Applied Linguistics*. Wiesbaden.

Waibel, A., Jain, A. N., McNair, A., Tebelskis, J., Osterholtz, L., Saito, H., Schmidbauer, O., Sloboda, T., & Woszczyna, M. (1992). JANUS: Speech-to-speech translation using connectionist and non-connectionist techniques. In Moody, J. E., Hanson, S. J., & Lippmann, R. R. (Eds.), *Advances in Neural Information Processing Systems 4*, pp. 183–190, Morgan Kaufmann, San Mateo, CA.

Ward, N. (1994). An approach to tightly-coupled syntactic/semantic processing for speech understanding. In *Proceedings of the 12th National Conference on Artificial Intelligence Workshop on the Integration of Natural Language and Speech Processing*, pp. 50–57, Seattle, Washington.

Weber, V., & Wermter, S. (1995). Towards learning semantics of spontaneous dialog utterances in a hybrid framework. In Hallam, J. (Ed.), *Hybrid Problems, Hybrid Solutions — Proceedings of the 10th Biennial Conference on AI and Cognitive Science*, pp. 229–238, Sheffield, UK.

Weber, V., & Wermter, S. (1996). Artificial neural networks for repairing language. In *Proceedings of the 8th International Conference on Neural Networks and their Applications*, pp. 117–123, Marseille, FRA.

Wermter, S., & Weber, V. (1996a). Interactive spoken-language processing in a hybrid connectionist system. *IEEE Computer – Theme Issue on Interactive Natural Language Processing*, July, 65–74.

Wermter, S. (1994). Hybride symbolische und subsymbolische Verarbeitung am Beispiel der Sprachverarbeitung. In Duwe, I., Kurfeß, F., Paaß, G., & Vogel, S. (Eds.), *Herbstschule Konnektionismus und Neuronale Netze.* Gesellschaft für Mathematik und Datenverarbeitung (GMD), Sankt Augustin, FRG.

Wermter, S. (1995). *Hybrid Connectionist Natural Language Processing.* Chapman and Hall, Thompson International, London, UK.

Wermter, S., & Löchel, M. (1994). Connectionist learning of flat syntactic analysis for speech/language systems. In Marinaro, M., & Morasso, P. G. (Eds.), *Proceedings of the International Conference on Artificial Neural Networks*, Vol. 2, pp. 941–944, Sorrento, Italy.

Wermter, S., & Löchel, M. (1996). Learning dialog act processing. In *Proceedings of the 16th International Conference on Computational Linguistics*, pp. 740–745, Kopenhagen, Denmark.

Wermter, S., & Peters, U. (1994). Learning incremental case assignment based on modular connectionist knowledge sources. In Werbos, P., Szu, H., & Widrow, B. (Eds.), *Proceedings of the World Congress on Neural Networks*, Vol. 4, pp. 538–543, San Diego, CA.

Wermter, S., Riloff, E., & Scheler, G. (Eds.). (1996). *Connectionist, Statistical and Symbolic Approaches to Learning for Natural Language Processing.* Springer, Berlin.

Wermter, S., & Weber, V. (1996b). Artificial neural networks for automatic knowledge acquisition in multiple real–world language domains. In *Proceedings of the 8th International Conference on Neural Networks and their Applications*, pp. 289–296, Marseille, FRA.

Wooters, C. C. (1993). Lexical modeling in a speaker independent speech understanding system. Tech. rep. TR-93-068, International Computer Science Institute, Berkeley.

Young, S. R., Hauptmann, A. G., Ward, W. H., Smith, E., & Werner, P. (1989). High level knowledge sources in usable speech recognition systems. *Communications of the ACM*, *32*, 183–194.